\documentclass[10pt,twocolumn,letterpaper]{article}

\usepackage{cvpr}              

\usepackage{subcaption}
\usepackage{graphicx}
\usepackage{amsmath}
\usepackage{amssymb}
\usepackage{booktabs}
\usepackage{verbatim}
\usepackage{multirow}
\usepackage{color}
\usepackage{bm}
\usepackage{indentfirst}

\usepackage[pagebackref,breaklinks,colorlinks]{hyperref}

\usepackage[capitalize]{cleveref}
\crefname{section}{Sec.}{Secs.}
\Crefname{section}{Section}{Sections}
\Crefname{table}{Table}{Tables}
\crefname{table}{Tab.}{Tabs.}

\def\cvprPaperID{5641} 
\def\confName{CVPR}
\def\confYear{2022}

\begin{document}

\title{DIP: Deep Inverse Patchmatch for High-Resolution Optical Flow}

\author{
    Zihua Zheng$^{1}$\ \ \ Ni Nie$^{1}$\ \ \ Zhi Ling$^{1}$\ \ \ Pengfei Xiong$^{2}$\ \ \  Jiangyu Liu$^{1}$\ \ \ Hao Wang$^{1}$\ \ \ Jiankun Li$^{1}$\ \ \ 
    \\
    $^{1}$Megvii \quad  $^{2}$Tencent \\
    \tt\small \{zhengzihua, nieni, lingzhi, liujiangyu, wanghao03, lijiankun\}@megvii.com \\ \tt\small xiongpengfei2019@gmail.com
    \vspace{-0.5em}
}

\maketitle

\begin{abstract}
Recently, the dense correlation volume method achieves state-of-the-art performance in optical flow. However, the correlation volume computation requires a lot of memory, which makes prediction difficult on high-resolution images. In this paper, we propose a novel Patchmatch-based framework to work on high-resolution optical flow estimation. 
Specifically, we introduce the first end-to-end Patchmatch based deep learning optical flow. It can get high-precision results with lower memory benefiting from propagation and local search of Patchmatch.
Furthermore, a new inverse propagation is proposed to decouple the complex operations of propagation, which can significantly reduce calculations in multiple iterations.
At the time of submission, our method ranks 1$^{st}$ on all the metrics on the popular KITTI2015\cite{geiger2013vision} benchmark
, and ranks 2$^{nd}$ on EPE on the Sintel\cite{butler2012naturalistic} clean benchmark
among published optical flow methods. Experiment shows our method has a strong cross-dataset generalization ability that the F1-all achieves 13.73$\%$, reducing 21$\%$ from the best published result 17.4$\%$ on KITTI2015.
What's more, our method shows a good details preserving result on the high-resolution dataset DAVIS\cite{almatrafi2019davis} and consumes 2$\times$ less memory than RAFT\cite{teed2020raft}.
Code will be available at \href{https://github.com/zihuazheng/DIP}{github.com/zihuazheng/DIP}
\end{abstract}

\section{Introduction}
\label{sec:intro}
\begin{figure}[htbp]
\centering
\begin{subfigure}[t]{0.23\textwidth}
    \centering
    \includegraphics[width=\textwidth]{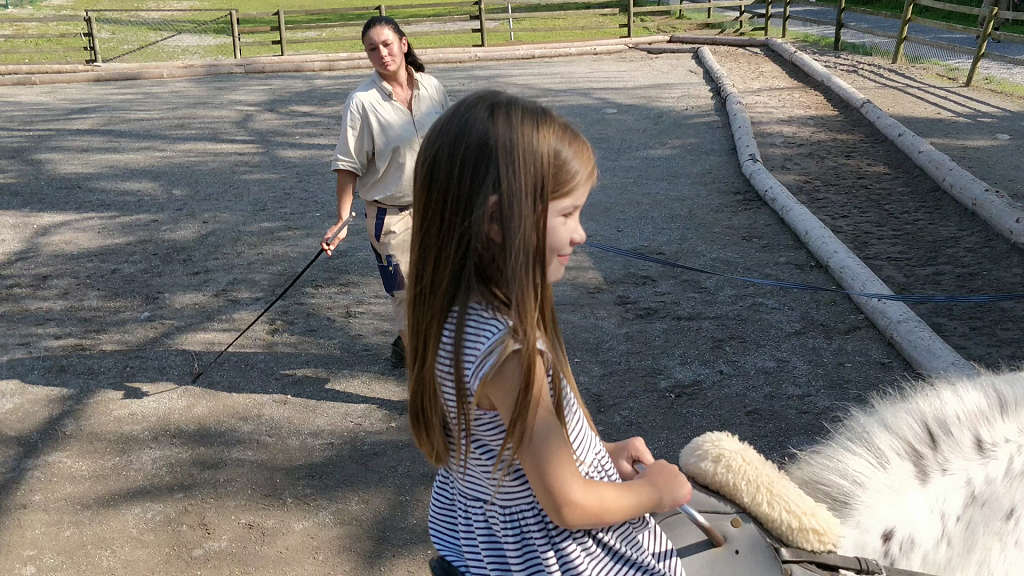}
    \caption{Image}
    \label{fig:Davis_reference}
\end{subfigure}
\begin{subfigure}[t]{0.23\textwidth}
    \centering
    \includegraphics[width=\textwidth]{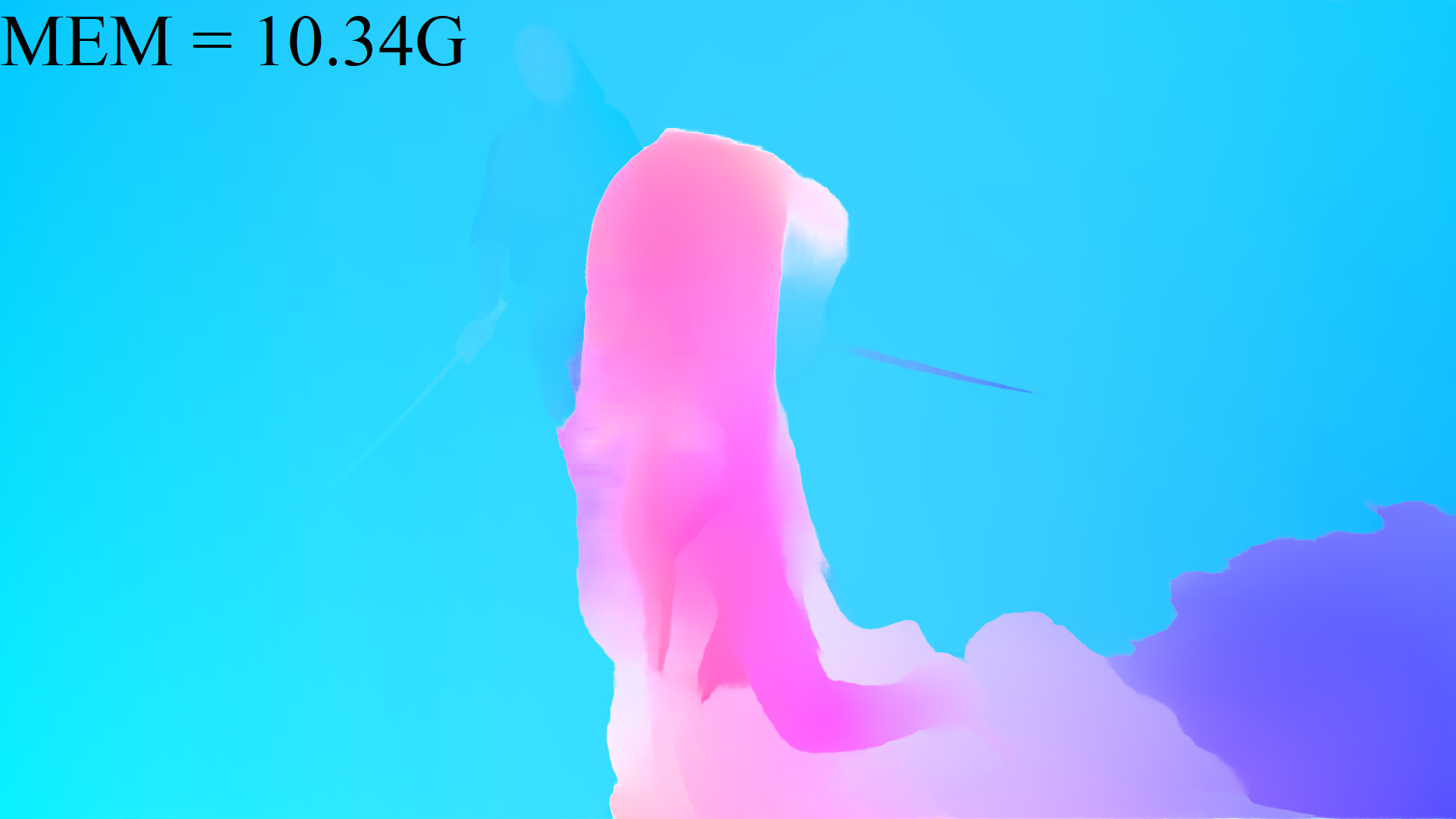}
    \caption{RAFT\cite{teed2020raft}}
    \label{fig:Davis_RAFT}
\end{subfigure}

\begin{subfigure}[t]{0.23\textwidth}
    \centering
    \includegraphics[width=\textwidth]{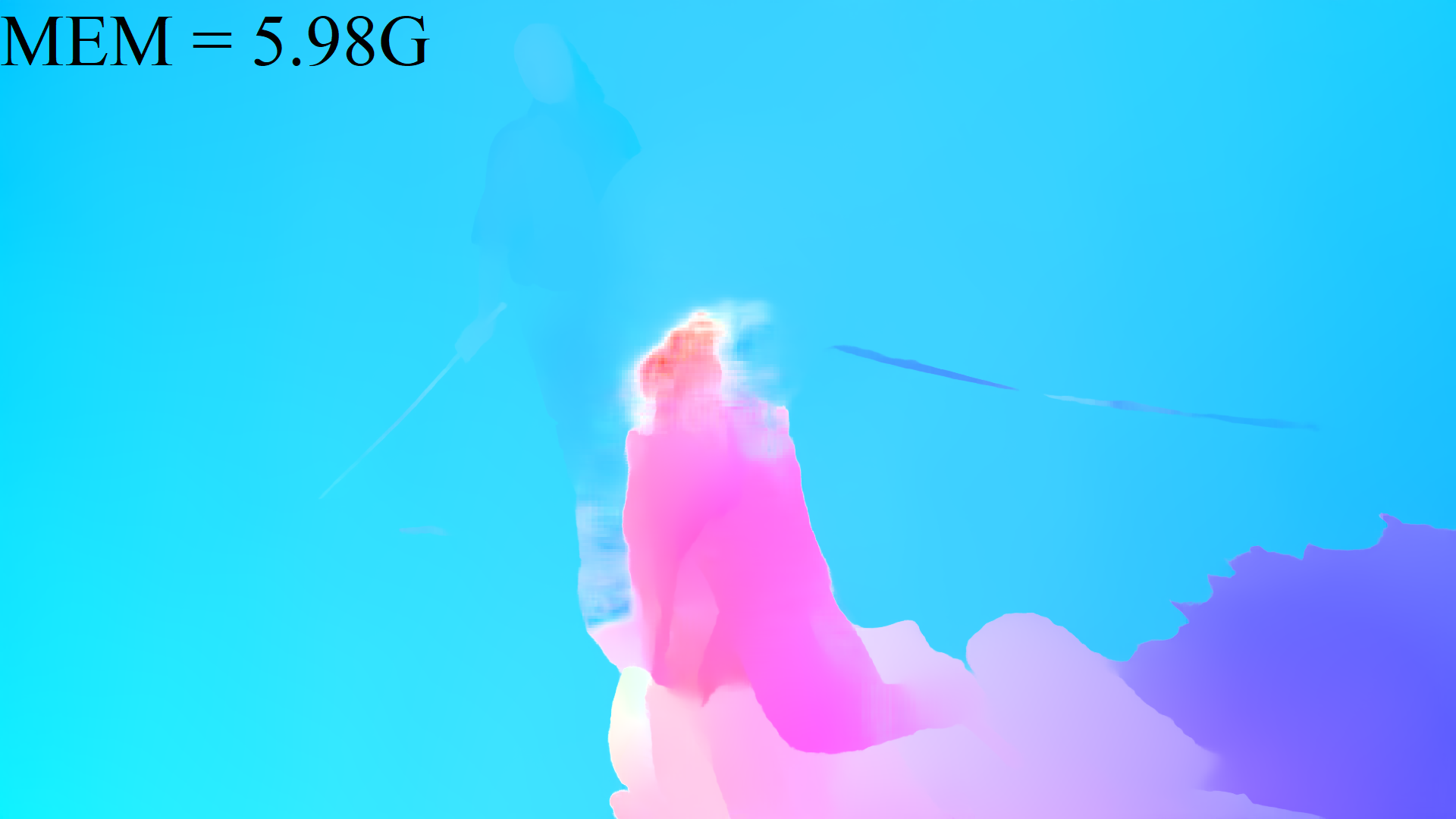}
    \caption{SCV\cite{jiang2021learning}}
    \label{fig:Davis_SCV}
\end{subfigure}
\begin{subfigure}[t]{0.23\textwidth}
    \centering
    \includegraphics[width=\textwidth]{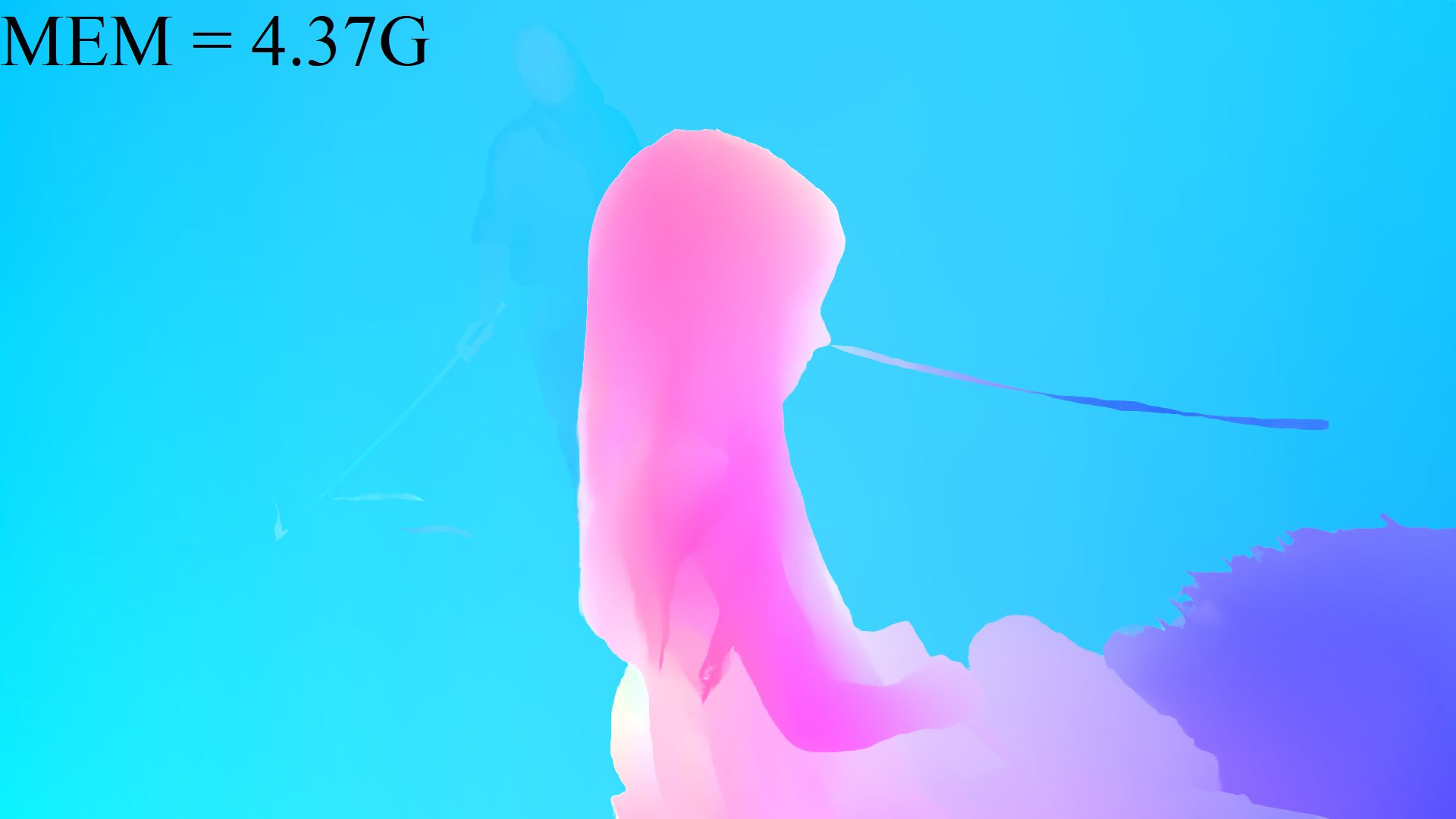}
    \caption{Ours}
    \label{fig:Davis_IPMFlow}
\end{subfigure}
\caption{Comparisons on high-resolution (1080 × 1920) images from DAVIS dataset. Compared with RAFT and SCV, our method has achieved better details with lower memory.}
\label{fig:result_compare}
\vspace{-0.2cm}
\end{figure}

Optical flow, the 2D displacement field that describes apparent motion of brightness patterns between two successive images \cite{horn1981determining}, provides valuable information about the spatial arrangement of the viewed objects and the change rate of the arrangement\cite{tu2019survey}. Since Horn and Schunck (HS) \cite{horn1981determining} and Lucas and Kanade (LK) \cite{lucas1981iterative} proposed the differential method to calculate optical flow in 1981, many extension algorithms\cite{weinzaepfel2013deepflow, kroeger2016fast, revaud2016deepmatching} have been proposed. Hence, optical flow has been widely used in various applications such as visual surveillance tasks\cite{xiao2016track}, segmentation\cite{tu2017fusing}, action recognition \cite{simonyan2014two}, obstacle detection \cite{ho2015optical} and image sequence super-resolution \cite{makansi2017end}.

Recently, deep learning has made great progress in solving the problem of optical flow. Since FlowNetC\cite{dosovitskiy2015flownet}, many methods have achieved state-of-the-art results. For deep learning, in addition to accuracy, performance and memory are also challenges especially when predicting flow at high-resolution. 
To reduce complexity of computation and usage of memory, previous approaches\cite{ilg2017flownet,hui2018liteflownet,sun2018pwc,yin2019hierarchical,hui2020lightweight} use coarse-to-fine strategy, they may suffer from low-resolution error recovery problems. In order to maintain high accuracy on large displacements, especially for fast moving small targets, RAFT\cite{teed2020raft} constructs an all-pairs 4D correlation volume and look up with a convolution GRU block. However, it runs into memory problems when predicting high-resolution optical flow.

In order to reduce the memory while maintaining high accuracy, instead of using the sparse global correlation strategies like \cite{jiang2021learning,xu2021high} which suffer from loss of accuracy, we introduce the idea of Patchmatch to the computation of correlation.
Patchmatch implements a random initialization, iterative propagation and search algorithm for approximate nearest neighbor field estimation\cite{barnes2009Patchmatch, bleyer2011patchmatch, hu2016efficient}. It only needs to perform correlation calculations on nearby pixels and propagate its cost information to the next matching point iteratively, without the need to construct a global matching cost. Therefore, the Patchmatch algorithm greatly reduces the memory overhead caused by the correlation volume. Moreover, the iterative propagation and search in Patchmatch can be easily achieved using GRU\cite{teed2020raft}.
To this end, we propose a Patchmatch-based framework for optical flow, which can effectively reduce memory while maintaining high accuracy. It contains two key modules: propagation module and local search module. The propagation module reduces the search radius effectively, and the local search module accelerates convergence and further improves accuracy. At the same time, we have achieved high-resolution predictions of high-precision optical flow through adaptive-layers iterations.

Furthermore, a new inverse propagation method is proposed, which offsets and stacks target patches in advance. Then, it only needs to do warping once for all propagations compared with propagation which requires offset and warping in each propagation, so as to reduce the calculation time significantly. 

We demonstrate our approach on the challenging Sintel\cite{butler2012naturalistic} and KITTI-15\cite{geiger2013vision} datasets. Our model ranks first on KITTI-15 and second on Sintel-Clean. \cref{fig:result_compare} shows the results of our Deep Inverse Patchmatch(DIP). Comparing to previous approaches \cite{teed2020raft, jiang2021learningSCV}, DIP keeps the best effect while memory usage is the lowest. 
At the same time, our method has a strong cross-dataset generalization that the F1-all achieves 13.73$\%$, reduced 21$\%$ from the best published result 17.4$\%$ on KITTI2015\cite{geiger2013vision}. In addition, the supplementary material shows the domain invariance of our DIP in the Stereo field.

To sum up, our main contributions include:
\vspace{-5pt}
\begin{itemize}
\item We design an efficient framework which introduces Patchmatch to the end-to-end optical flow prediction for the first time. It can improve the accuracy of optical flow while reducing the memory of correlation volume.
\vspace{-5pt}
\item We propose a novel inverse propagation module. Compared with propagation, it can effectively reduce calculations while maintaining considerable performance. 
\vspace{-5pt}
\item Our experiments demonstrate that the method achieves a good trade-off between performance and memory, a comparable results with the state of the art methods on public datasets and a good generalization on different datasets.
\end{itemize}

\section{Related Work}
\label{sec:Related_Work}
\paragraph{Deep Flow Methods} The first end-to-end CNN-based version for flow estimation can be traced back to \cite{dosovitskiy2015flownet}, which proposed a U-net like architecture FlowNetS to predict flow directly. A correlation layer was included in a diverse version named FlowNetC. In FlowNet2, Ilg \etal. \cite{ilg2017flownet} introduced a warping mechanism and stacked hourglass network to promote the performance on small motion areas. PWC-Net \cite{sun2018pwc} used feature warping and a coarse-to-fine cost volume with a context network for flow refinement, further improving the accuracy and reducing the model size simultaneously. To address ambiguous correspondence and occlusion problem, Hui \etal. \cite{hui2020liteflownet3} proposed LiteFlowNet3 with adaptive affine transformation and local flow consistency restrictions. RAFT \cite{teed2020raft} introduced a shared weight iterative refinement module to update the flow field retrieved from a 4D all-pair correlation volume. To reduce the computation complexity of 2D searching in high-resolution images, Xu \etal. \cite{xu2021high} factorized the 2D search to 1D in two directions combined with attention mechanism. Jiang \etal. \cite{jiang2021learningSCV} proposed to construct a sparse correlation volume directly by computing the k-Nearest matches in one feature map for each feature vector in the other feature map. The memory consumption of them is less compare to RAFT but their accuracy is inferior. Another line of work is focused on joining image segmentation and flow estimation task together \cite{chang2013topology, sun2013fully, tsai2016video, cheng2017segflow}, which propagated two different complementary features, aiming at improving the performance of flow estimation and vice versa.
\vspace{-10pt}
\paragraph{Patchmatch Based Methods} Patchmatch has been originally proposed by Barnes \etal. \cite{barnes2009Patchmatch}. Its core work is to compute patch correspondences in a pair of images. The key idea behind it is that neighboring pixels usually have coherent matches. M Bleyer \etal.\cite{bleyer2011patchmatch} applied Patchmatch to stereo matching and proposed a slanted support windows method for computing aggregation to obtain sub-pixel disparity precision. In order to reduce the error caused by the motion discontinuity of Patchmatch in optical flow, Bao \etal.\cite{bao2014fast} proposed the Edge-Preserving Patchmatch algorithm. Hu \etal.\cite{hu2016efficient} proposed a Coarse-to-Fine Patchmatch strategy to improve the speed and accuracy of optical flow. In deep learning, Bailer \etal. \cite{bailer2017cnn} regarded Patchmatch as a 2-classification problem and proposed a thresholded loss to improve the accuracy of classification.
Shivam \etal.\cite{duggal2019deeppruner} developed a differentiable Patchmatch module to achieve real-time in the stereo disparity estimation network. But this method is sparse and only works on the disparity dimension. Wang \etal.\cite{wang2021patchmatchnet} introduced iterative multi-scale Patchmatch, which used one adaptive propagation and differentiable warping strategy, achieved a good performance in the Multi-View Stereo problem.

\section{Method}
\begin{figure*}[htbp]
\centering
\begin{subfigure}[t]{0.48\textwidth}
\centering
\includegraphics[width=8cm]{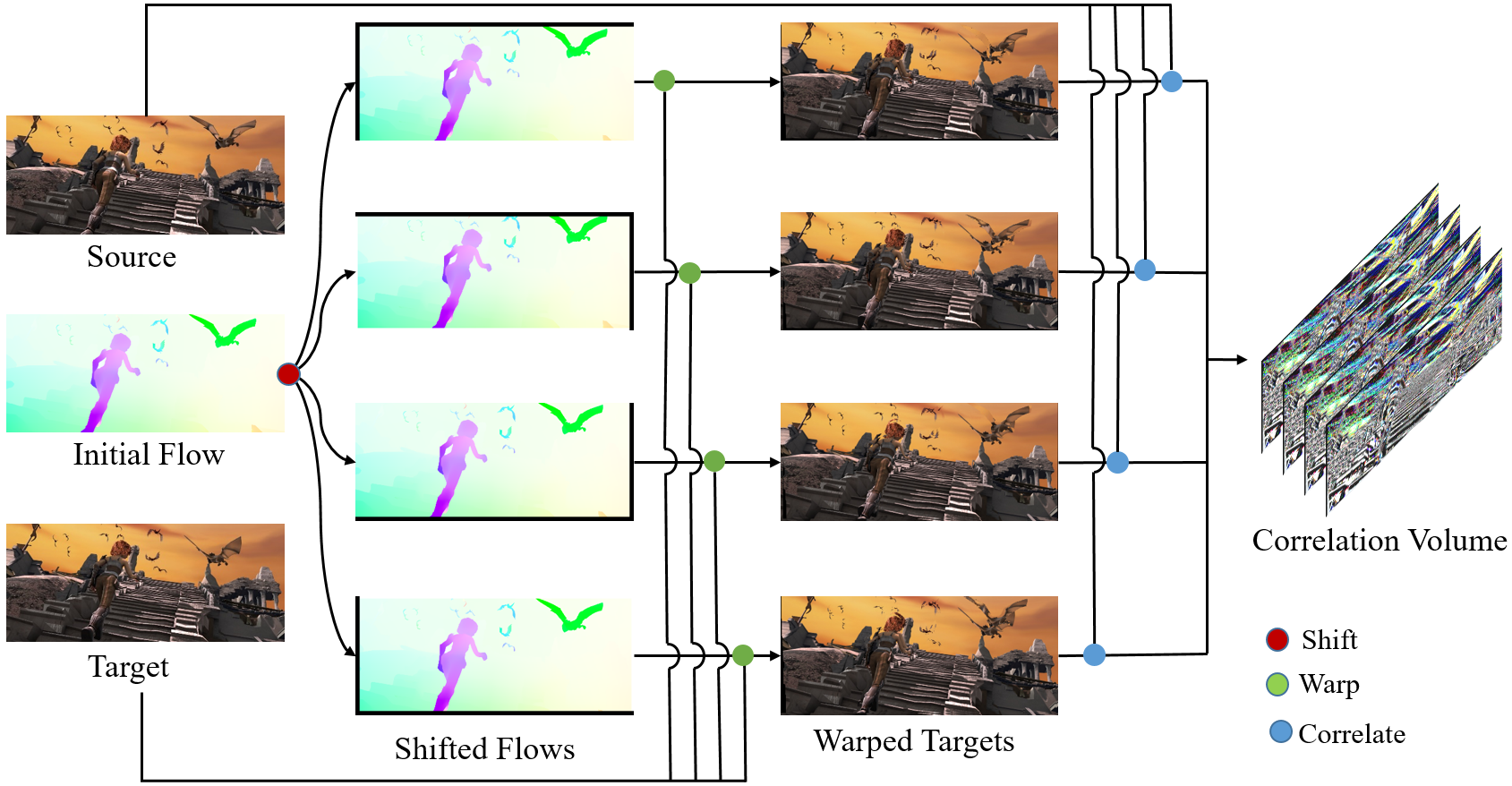}
\caption{propagation}
\label{fig:deep_propagation}
\end{subfigure}
\begin{subfigure}[t]{0.48\textwidth}
\centering
\includegraphics[width=8cm]{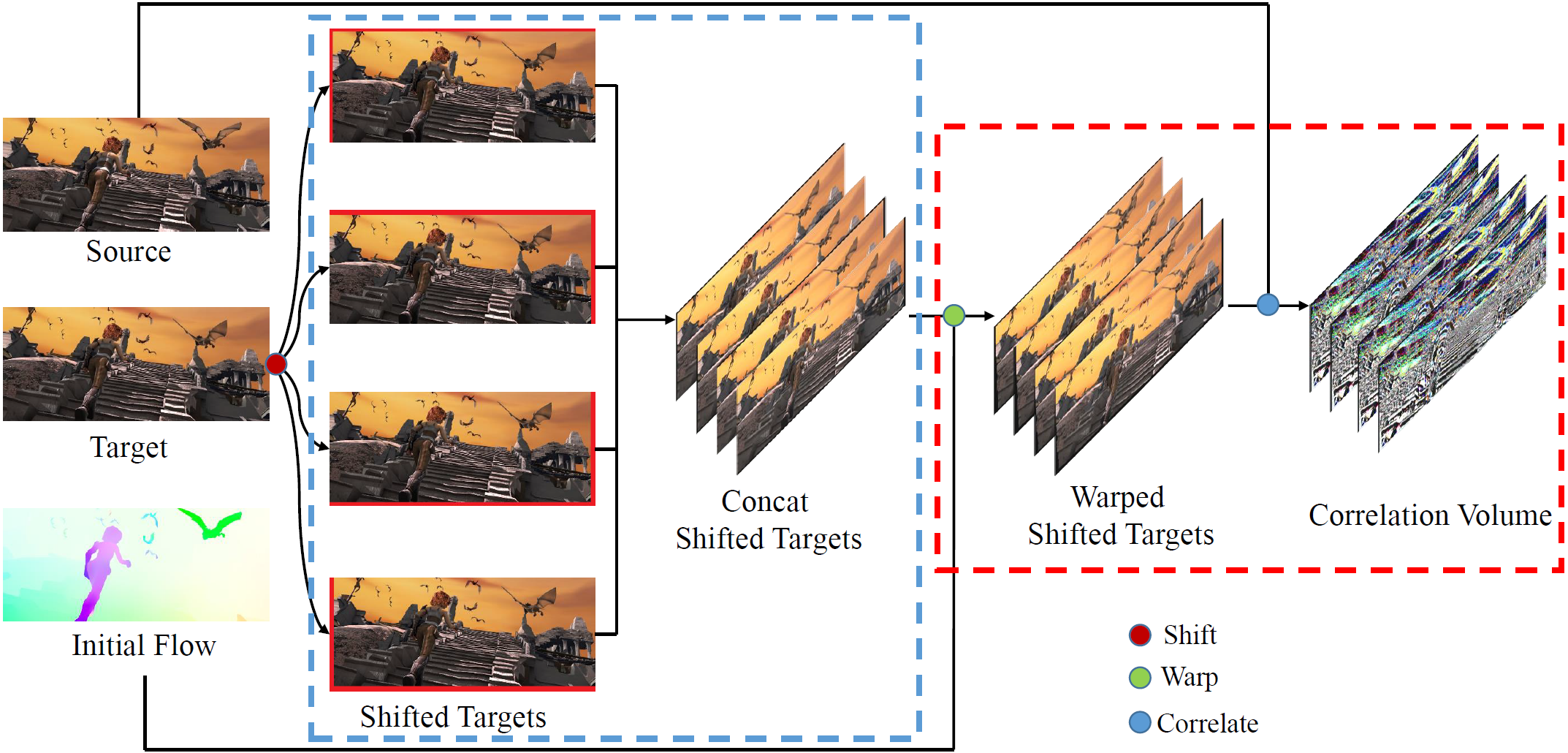}
\caption{inverse propagation}
\label{fig:deep_inverse_propagation}
\end{subfigure}
\caption{The correlation calculation process of propagation and inverse propagation. Where the red points in the graph represent shift operator on the optical flow or images according to the seed points, the green points represent warping operator on the images according to the optical flow, and the blue points represent correlation caculating operator between the source image and the warped images. The blue box in (b) represents the initialization stage, and the red box represents the running stage.}
\label{fig:short}
\end{figure*}

\label{sec:Approach}
We start with our observation and analysis of different correlation volume in optical flow task. These methods require high memory usage and computation to compute the correlation volume. Inspired by the high efficiency of Patchmatch on the correspondence points matching, we use it to reduce the search space of optical flow.

\subsection{Observations}
\label{sec:Observations}

\paragraph{Local Correlation Volume}
In modern local correlation volume based optical flow approaches\cite{dosovitskiy2015flownet}, the computation of it can be formulated as follows:
\begin{equation}
    Corr = \left \{ F_1(\mathbf{x}) \cdot F_2(\mathbf{x}+\mathbf{d})|\mathbf{x}\in X,\mathbf{d}\in D \right \},
    \label{eq:common_corr}
\end{equation}
where $F_1$ is the source feature map and $F_2$ is the target feature map, $\mathbf{d}$ is the displacement along the $x$ or $y$ direction. $X = [0, h)\times[0, w)$, $D = [-d_{max}, d_{max}]^2$, $h$ is the height of feature map, $w$ is the width of feature map. So the memory and calculation of the correlation volume are linear to $hw(2d_{max}+1)^2$ and quadratic to the radius of the search space. Limited by the size of the search radius, it is difficult to obtain high-precision optical flow in high-resolution challenging scenes.
\vspace{-10pt}
\paragraph{Global Correlation Volume}
Recently, RAFT\cite{teed2020raft} proposed an all-pairs correlation volume which achieved the state-of-the-art performance. The global correlation computation at location ($i$, $j$) in $F_1$ and location ($k$, $l$) in $F_2$ can be defined as follows:
\begin{equation}
Corr_{i j k l}^{m}=\frac{1}{2^{2 m}} \sum_{p}^{2^{m}} \sum_{q}^{2^{m}}\left( F_{1}(i,j) \cdot F_{2}(2^{m}k+p, 2^{m}l+q)\right),
\end{equation}
where $m$ is the pyramid layer number. $2^{m}$ is the pooled kernel size. 
Compared with local correlation volume, global correlation volume contains $N^2$ elements, where $N$ = $hw$. When the $h$ or $w$ of $F$ increases, the memory and calculation will multiply. So the global method suffers from insufficient memory when inferring at high- resolution. 
\vspace{-10pt}
\paragraph{Patchmatch Method} Patchmatch is proposed by Barnes \etal. \cite{barnes2009Patchmatch} to find dense correspondences across images for structural editing. The key idea behind it is that we can get some good guesses by a large number of random samples. And based on the locality of image, once a good match is found, the information can be efficiently propagated to its neighbors. 
So, we propose to use the propagation strategy to reduce the search radius and use local search to further improve accuracy. And the complexity of Patchmatch method is $hw(n + r^2)$, where $n$ is the number of propagation, $r$ is the local search radius, and both values are very small and do not change with the increase of displacement or resolution.
Details are described in the next subsection.

\subsection{Patchmatch In Flow Problem}
\label{sec:Traditional_Patchmatch}
\begin{figure*}[ht]
\centering
\begin{subfigure}[t]{1.0\textwidth}
\centering
\includegraphics[width=16cm]{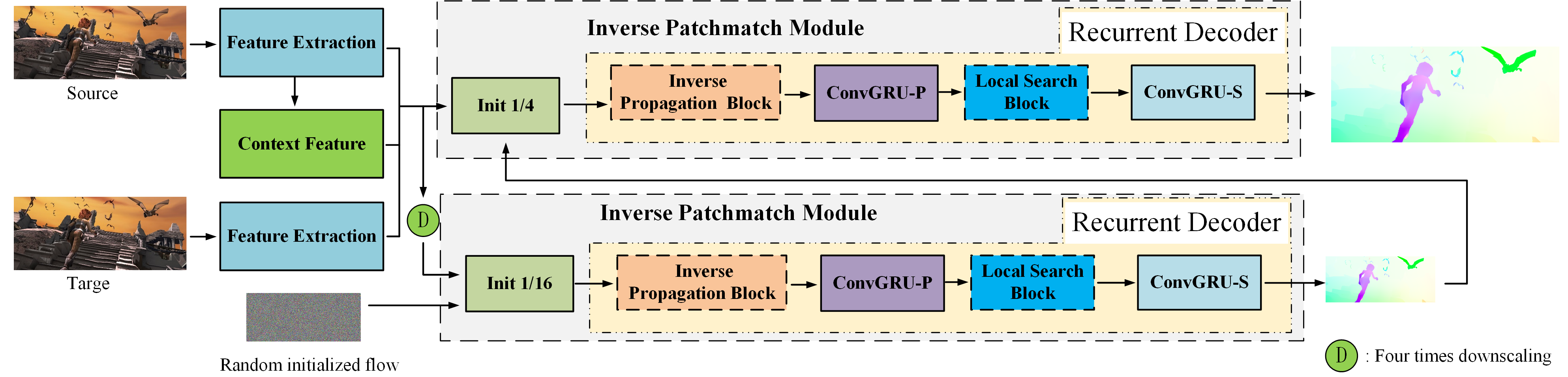}
\caption{DIP Architecture}
\label{fig:total_pipeline_sub}
\end{subfigure}

\begin{subfigure}[t]{0.54\textwidth}
\centering
\includegraphics[width=9.25cm]{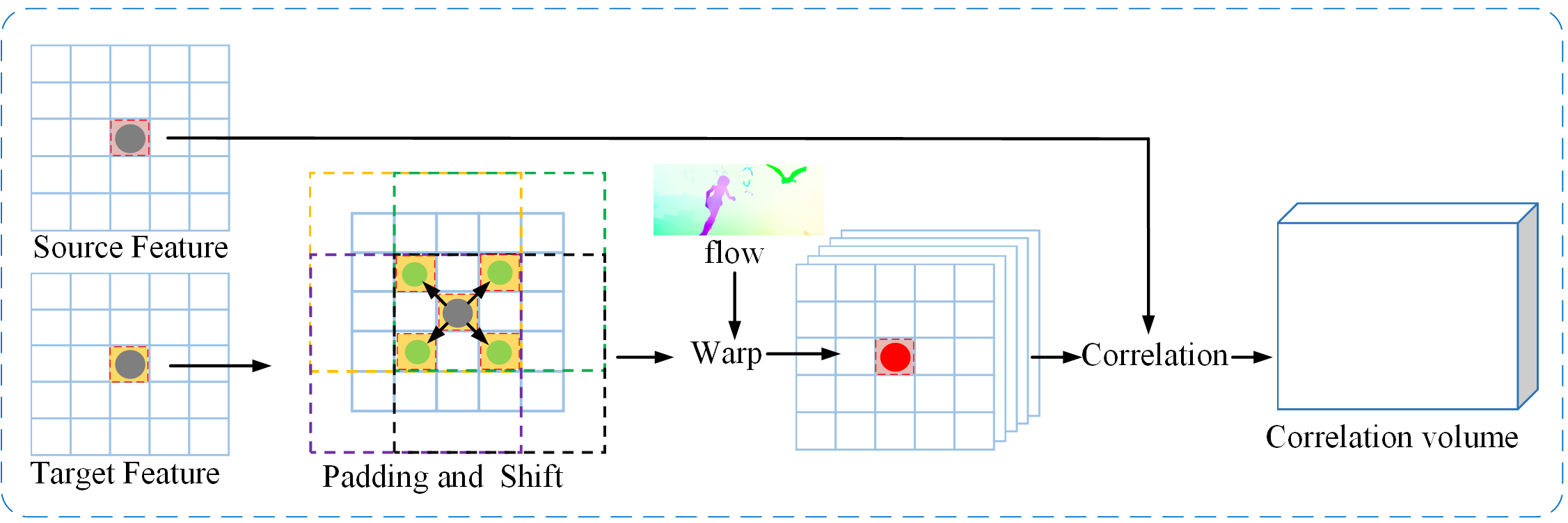}
\caption{Inverse Propagation Block}
\label{fig:inverse_propagation_block}
\end{subfigure}
\begin{subfigure}[t]{0.44\textwidth}
\centering
\includegraphics[width=7.4cm]{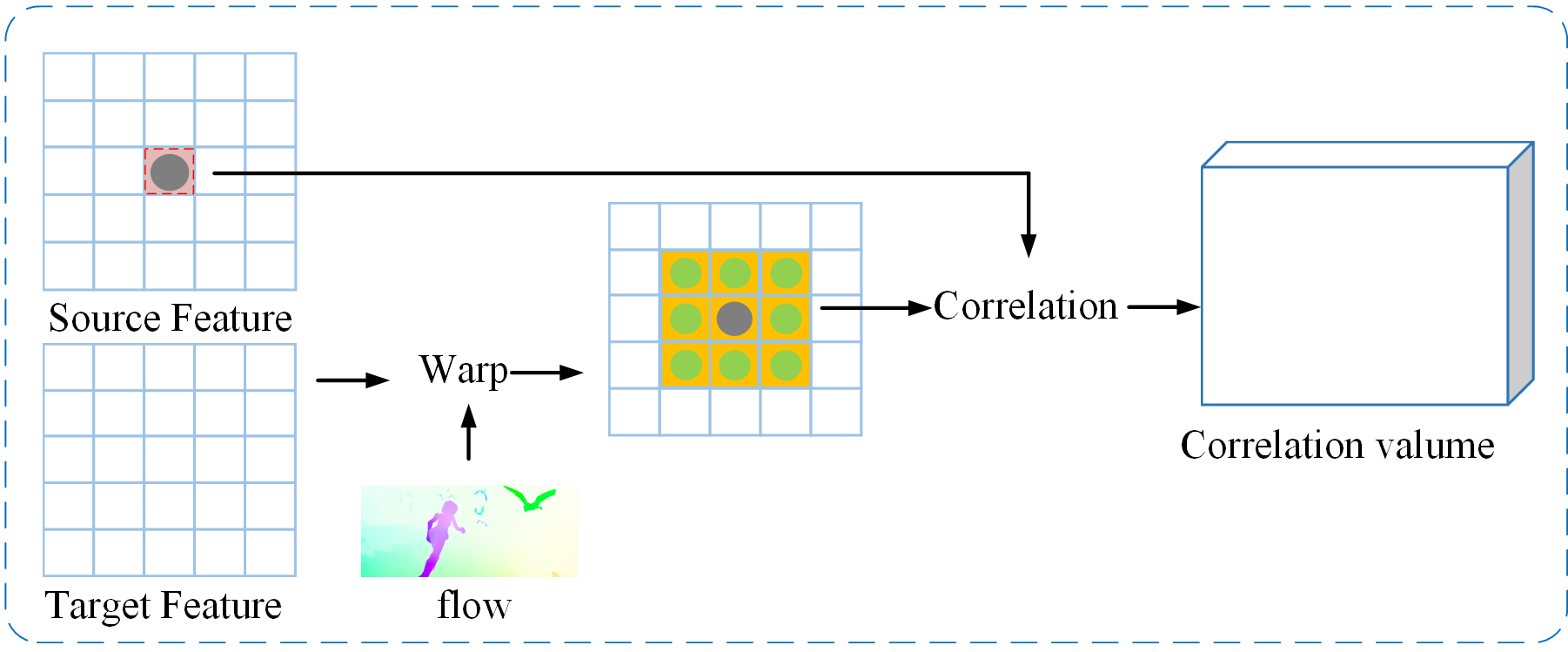}
\caption{Local Search Block}
\label{fig:local_search_block}
\end{subfigure}
\caption{(a) Architecture overview. Given a pair of flow images, we first extract deep 1/4 and 1/16 scale features and context information. The extracted features and context information are then used to the initialization of 1/16 inverse Patchmatch, which is fed into the inverse propagation Block, local search Block and GRU modules for iterative optimization of flow. Then we use the optical flow predicted on 1/16 for the initialization of 1/4 inverse Patchmatch, and repeat the operation of Inverse Patchmatch Network. Please note that the parameters used by 1/4 and 1/16 Inverse Patchmatch Network are exactly the same. (b) Inverse Propagation Block propagates neighbor information. (c) Local Search Block is used to refine the flow. }
\label{fig:total_pipeline}
\vspace{-3mm}
\end{figure*}

The traditional Patchmatch methods\cite{barnes2009Patchmatch,bleyer2011patchmatch,hu2016efficient,kuang2017patchmatch}
has three main components. 1) Random initialization. It gets some good guesses by a large number of random samples. 2) Propagation. Based on image locality, once a good match is found, the information can be efficiently propagated from its neighbors. 3) Random search. It is used in the subsequent
propagation to prevent local optimization and make it possible to obtain the good match when no good match exist in its neighbors.

Iterative propagation and search are the key points to solve the flow problem.
In propagation stage, we treat a point of feature maps as a patch and select 4 neighbor seed points. So every point can get the flow candidates from its neighbors by shifting the flow map toward the 4 neighbors. Then we can compute a 5 dimension correlation volume based on the neighbor flow candidates and its flow. Given a shift $\Delta p$ for all flow, the correlation calculation of propagation can be defined as:
\begin{equation}
    Corr =   F_{1}\cdot \textbf{W}(F_{2}, \textbf{S}(flow, \Delta p) ),
    \label{eq:patch_match}
\end{equation}
Where, ${\textbf{S}(flow, \Delta p)}$ refers to shift flow according to ${ \Delta p}$, $\textbf{W}$ refers to warp $F_{2}$ with shifted flow.  There is no doubt that the more seed points are selected, the more operations are needed. When choosing $n$ seed points for $m$ iterations of propagation, propagation needs to shift the optical flow $n\times m$ times and warp the source feature $n\times m$ times. This increases memory operations and interpolation calculations, especially when predicting high-resolution optical flow. In order to reduce the number of options, for the first time we replace propagation with inverse propagation. In the search stage, we change the random search to a local search method which is more suitable for end-to-end network and achieves higher accuracy. More details in patchmatch method can be seen in the supplementary.
\subsection{Deep Inverse Patchmatch}
\label{sec:Deep_Inverse_Patch_Patch}
\noindent{\bf Inverse Propagation}
In propagation, the optical flow shift and feature warping are serial and coupled, since the warping process depends on the shifted flow. Moreover, multiple flow shifts are necessary in each iteration, so the computations increase. In theory, the spatial relative position of shifting the flow to the down-right is the same as shifting the target to the top-left. And the correlation maps of the two methods have one pixel offset in the absolute space coordinates. We name the way of shifting targets as inverse propagation,
and the inverse propagation can be formulated as follows:
\begin{small}
\begin{equation}
    Corr = F_{1}\cdot \textbf{S}(F_{2}^{'}, -\Delta p),
    \label{eq:inverse_patch_match_step2}
\end{equation}
and
\begin{equation}
    F_{2}^{'}= \textbf{W}(\textbf{S}(F_{2}, \Delta p), flow)
    \label{eq:inverse_patch_match_step1}
\end{equation}
\end{small}

In theory, combining \cref{eq:inverse_patch_match_step1} and \cref{eq:inverse_patch_match_step2} is completely equivalent to \cref{eq:patch_match}. Since $\Delta p$ is very small, we ignore the process of back propagation in our implementation. Then \cref{eq:inverse_patch_match_step2} can be replaced with:
\begin{equation}
    Corr = F_{1}\cdot F_{2}^{'}
    \label{eq:inverse_patch_match_step2_new}
\end{equation}
In inverse propagation, a target feature point is scattered to its seed points and warped by the optical flow of the seed points. Thus, we can shift and stack the target features in advance, then perform warping only once to obtain the warped target features in each iteration. The details of inverse propagation can be described in \cref{fig:inverse_propagation_block}.

In this work, the seed points is static and do not change with the increase of iterations. Hence target features only need to be shifted to seed points once and shifted target features can be reused in every iteration. In this way, if there are $n$ seed points for $m$ iterations of propagation, we only need to shift target features $n$ times and warp the shifted target features $m$ times. \cref{fig:deep_inverse_propagation} shows the inverse propagation stage and whole the stage can be divided into two sub-stages:
\begin{itemize}
\item \textbf{Initialization Stage:} Input source feature, target feature. Shift the target feature according to the seed points, and then stack these shifted target features as shared target features along the depth dimension. 
\item \textbf{Running Stage:} Input a flow, warp shared target features according to the flow, and compute correlation between source feature and warped target features.
\vspace{-10pt}
\end{itemize}
\paragraph{Local Search}
It is difficult to obtain very accurate optical flow by patch propagation alone, since the range of randomly initialized flow values is very sparse. Therefore, a local neighborhood search is performed after each patch propagation in this work. Unlike \cite{barnes2009Patchmatch}, which performs a random search after each propagation and reduces the search radius with increasing iteration. We only perform a fixed small radius search after each propagation and call it local search. The entire local search block is shown in \cref{fig:local_search_block}. Given an optical flow increment $\Delta f$, the local search can be formulated as:
\begin{equation}
    Corr = F_{1}\cdot \textbf{S}(\textbf{W}(F_{2}, flow), \Delta f)
    \label{eq:forward_search}
\end{equation}
In this work, we set the final search radius to 2 according to the experimental results. Details are described in \Cref{sec:Ablation_study}.

To this end, the Inverse Patchmatch module, as shown in \cref{fig:total_pipeline_sub}, consists mainly of the Inverse Propagation Block and the Local Search Block. In each iteration, an inverse propagation is followed by a local search. It is worth noting that both blocks use GRU\cite{teed2020raft} for cost aggregation.
\subsection{Network Architecture}
\label{sec:Network_Architecture}
In order to obtain high-precision optical flow on high-resolution images, we designed a new optical flow prediction framework named DIP. The overview of DIP can be found in \cref{fig:total_pipeline}. It can be described as two main stages: (1) feature extraction; (2)multi-scale iterative update. 
\vspace{-8pt}
\paragraph{Feature Extraction}
At first, a feature encoder network is applied to the input images to extract the feature maps at 1/4 resolution. Unlike previous works \cite{teed2020raft,jiang2021learningSCV,xu2021high,jiang2021learning} which use a context network branch to specifically extract the context. DIP directly activates the source feature map as a context map. Then we use the Average Pooling module to reduce the feature maps to 1/16 resolution. And we use the same backbone and parameters for both 1/4 resolution and 1/16 resolution. Therefore, DIP can be trained in two stages, and we use more stages for inference when processing large images.
\vspace{-8pt}
\paragraph{Multi-scale Iterative Update}
Our method is based on neighborhood propagation and thus must iteratively update the optical flow. Our network consists of two modules, an inverse propagation module and a local search module. In the training stage, we start the network with a random flow of size 1/16 and then iteratively optimize the optical flow at both scale 1/16 and scale 1/4 using a pyramid method. During the inference stage, we can perform the same process as in the training stage. To obtain a more accurate optical flow, we can also refine the optical flow at scale 1/8 and then optimize the result at scale 1/4. More high-resolution detailed comparisons can be found in the supplementary material.
 
Our network also accepts the initialized optical flow as input in the inference stage. In this case, we adapt the number of inference layers of the pyramid according to the maximum value of the initialized optical flow. For example, the forward interpolation of the optical flow of the previous image is used as input for the current image when the optical flow of the video images is processed. With the information of the previous optical flow, we can use two or more pyramids for large displacements to ensure accuracy, and use one pyramid for small displacements to reduce inference time.

\section{Experiment}
\begin{table*}[htbp]
  \centering
    \begin{tabular}{cccccccccc}
    \toprule
     \multirow{2}{*}{Method} & \multicolumn{2}{c}{Sintel (train)} & \multicolumn{2}{c}{KITTI-15 (train)} &  \multirow{2}[4]{*}{Params} & \multicolumn{2}{c}{448$\times$1024} & \multicolumn{2}{c}{1088$\times$1920} \\
     \cmidrule(lr){2-3} \cmidrule(lr){4-5}  \cmidrule(lr){7-8}\cmidrule(lr){9-10}
      & Clean & Final & EPE   & F1-all &    & Memory & Time(ms) & Memory & Time (ms) \\
    \midrule
    Sparse global\cite{jiang2021learningSCV} &\underline{1.29}  & \underline{2.95}  & 6.80   &19.30    &5.00M &3.04G &839   &5.98G   &3971 \\
    Dense global & 1.30  & 2.97  &\underline{4.96}     &\textbf{14.02}    & 3.40M & 10.47G & 234   & OOM   & - \\
    \midrule
    only p(N=4) & 1.62  & 3.40   & 7.63   & 19.81  & 2.78M   & \textbf{1.48G}    & 112   & \textbf{3.27G}   & \textbf{325} \\
    only ls(r=1) & 1.48  &  3.02  &12.38   &  23.76    & 3.40M   &  1.56G   &\textbf{96}     & \underline{3.45G}  & 373 \\
    pm(N=4, r=1) &\textbf{1.26}  &\textbf{2.93}  &\textbf{4.89}  & \underline{14.33} & 5.10M  & \underline{1.56G}  & \underline{106} & 3.70G  & \underline{372} \\
    \bottomrule
    \end{tabular}
  \caption{Ablation study concerning correlation volume. Models are trained on FlyingChairs\cite{dosovitskiy2015flownet} and FlyingThings3D\cite{mayer2016large}. Memory and inference time are measured on a RTX2080 Ti GPU. \textit{global} means global correlation volume. \textit{only p(N=4), ls(r=1)} means that only use propagation with seeds of 4 or local search with radius 1. \textit{pm(N=4, r=1)} means Patchmatch that combines propagation and local search. The number of iterations is set to 6 for Patchmatch and 12 for other methods. The best results are marked with bold and the second best results are marked with underline.}
  \label{tab:global_local}
  \vspace{-4mm}
\end{table*}

\begin{table}[htbp]
  \centering
    \begin{tabular}{ccccccc}
    \toprule
    \multicolumn{2}{c}{pm} & \multicolumn{2}{c}{Sintel} & \multicolumn{2}{c}{KITTI-15}  & 1088$\times$1920 \\
    \cmidrule(lr){1-2} \cmidrule(lr){3-4} \cmidrule(lr){5-6}
    $N$  &$r$      & \multicolumn{1}{c}{clean} & \multicolumn{1}{c}{final} & \multicolumn{1}{c}{EPE} & \multicolumn{1}{c}{F1-all} &Time(ms) \\
    \midrule
    4  &1 &\textbf{1.26} & 2.93 & 4.89 & 14.33 &\textbf{372}\\
    \textbf{4}  &\textbf{2} &\underline{1.27} & \underline{2.83} & \textbf{4.41} & \textbf{13.51} &\underline{432}\\
    4  &3 &1.31 & 2.85 & 4.54 & 13.80 &523\\ 
    8  &2 &1.28 & \textbf{2.79} & \underline{4.45} & \underline{13.77} &503\\
    \bottomrule
    \end{tabular}%
  \caption{Ablation study of the number of seeds and the local search radius based on Patchmatch. Validated on Sintel and KITTI-15 training datasets and iteration is set to 6. The best results are marked with bold and the second best results are marked with underline.}
  \label{tab:compare_pm_ls}%
\end{table}%

\begin{table}[htbp]
  \centering
    \begin{tabular}{cccccc}
    \toprule
    \multirow{2}{*}{Method} & \multicolumn{2}{c}{Sintel} & \multicolumn{2}{c}{KITTI-15} & 1088$\times$1920 \\
    \cmidrule(lr){2-3} \cmidrule(lr){4-5}
    &clean &final &EPE   &F1-all  &Time(ms) \\
    \midrule
    pm & \textbf{1.27}  & 2.83  & 4.41  & \textbf{13.51} & 432 \\
    \textbf{ipm} & 1.30  & \textbf{2.82}  & \textbf{4.29}  &13.73 & \textbf{327} \\
    \bottomrule
    \end{tabular}%
    \caption{Ablation study of Patchmatch and inverse Patchmatch on Sintel and KITTI-15 training datasets. Where $pm$ means Patchmatch and $ipm$ means inverse Patchmatch. Among them, the seeds of propagation is 4. The radius of local search is 2. The best results are marked in bold.}
    \label{tab:pm_ipm}
\end{table}%

\label{sec:Experiments}
In this section we demonstrate the state-of-the-art performance of DIP on Sintel\cite{butler2012naturalistic} and KITTI\cite{geiger2013vision} leaderboards and show that it outperforms existing methods in the zero-shot generalization setting on Sintel and KITTI. The endpoint error (EPE) is reported in the evaluation. For KITTI, another evaluation metric, F1-all, is also reported, which indicates the percentage of outliers for all pixels. For benchmark performance evaluation, \emph{d}$_{0-10}$ and \emph{d}$_{10-60}$ on Sintel are also used to estimate the optical flow in small motion regions. Here, \emph{d}$_{0-10}$ means the endpoint error over regions closer than 10 pixels to the nearest occlusion boundary.

\subsection{Training schedule}
\label{sec:Training_schedule}
DIP is implemented in Pytorch \cite{paszke2019pytorch} with 16 RTX 2080 Ti GPUs. Following RAFT \cite{teed2020raft}, we use the AdamW \cite{loshchilov2017decoupled} optimizer and the OneCycle learning rate schedule \cite{smith2019super} in the training process.
\vspace{-8pt}
\paragraph{Training Details} In the generalization experiment, we train our model on the datasets FlyingChairs\cite{dosovitskiy2015flownet} and FlyingThings3D\cite{mayer2016large} and evaluate the generalization ability on the training set of Sintel \cite{butler2012naturalistic} and KITTI2015 \cite{geiger2013vision}. In the pre-train stage, we decide to combine FlyingChairs and FlyingThings3D in a ratio of 1:10. First, the training size is set to $512 \times384$, and the model is trained for 100k steps with a batch size of 32. Then the model is finetuned on size of $768 \times384$ for another 100k steps with batch size of 16. During training and inference of ablation studies, we use 6 iterations for DIP flow regression. And the number of iterations is set to 12 during benchmark performance evaluation.

We also performed fine-tuning on Sintel\cite{butler2012naturalistic}, KITTI\cite{geiger2013vision} and HD1K\cite{kondermann2016hci} datasets. We perform fine-tuning on Sintel for 100k by combining data from Sintel and FlyingThings3D\cite{mayer2016large} and training size is $768 \times384$. Finally, we perform fine-tuning using a combination of data from FlyingThings, Sintel, KITTI-15, and HD1K for 100k with a training size of $832 \times320$.
\vspace{-8pt}
\paragraph{Loss}
Our loss function is similar with RAFT \cite{teed2020raft}. DIP outputs two optical flows for each iteration. Thus, $N = iters \times 2 \times 2$ predictions are output throughout the training process when $N$ iterations are used at both 1/16 and 1/4 resolution. Since there are multiple outputs for supervise, we use the similar strategy with RAFT, to compute a weighting sequence and sum the loss of the prediction sequence with it. The total loss can be formulated as follows:
\begin{equation}
    loss = \sum_{i=0}^{i=N} w_i \cdot M( \left | f_i-f_{gt} \right | ),
    \label{eq:loss}
\end{equation}

where $N$ is the length of the prediction sequence, $M(x)$ represents the mean of the matrix $x$, and the $w_i$ can be computed by \cref{eq:weight}, we use $\gamma=0.8$ in our training.

\begin{equation}
    w_i = \gamma^{N - i - 1}
    \label{eq:weight}
\end{equation}

\begin{figure*}[htbp]
\centering 
\includegraphics[width=17cm]{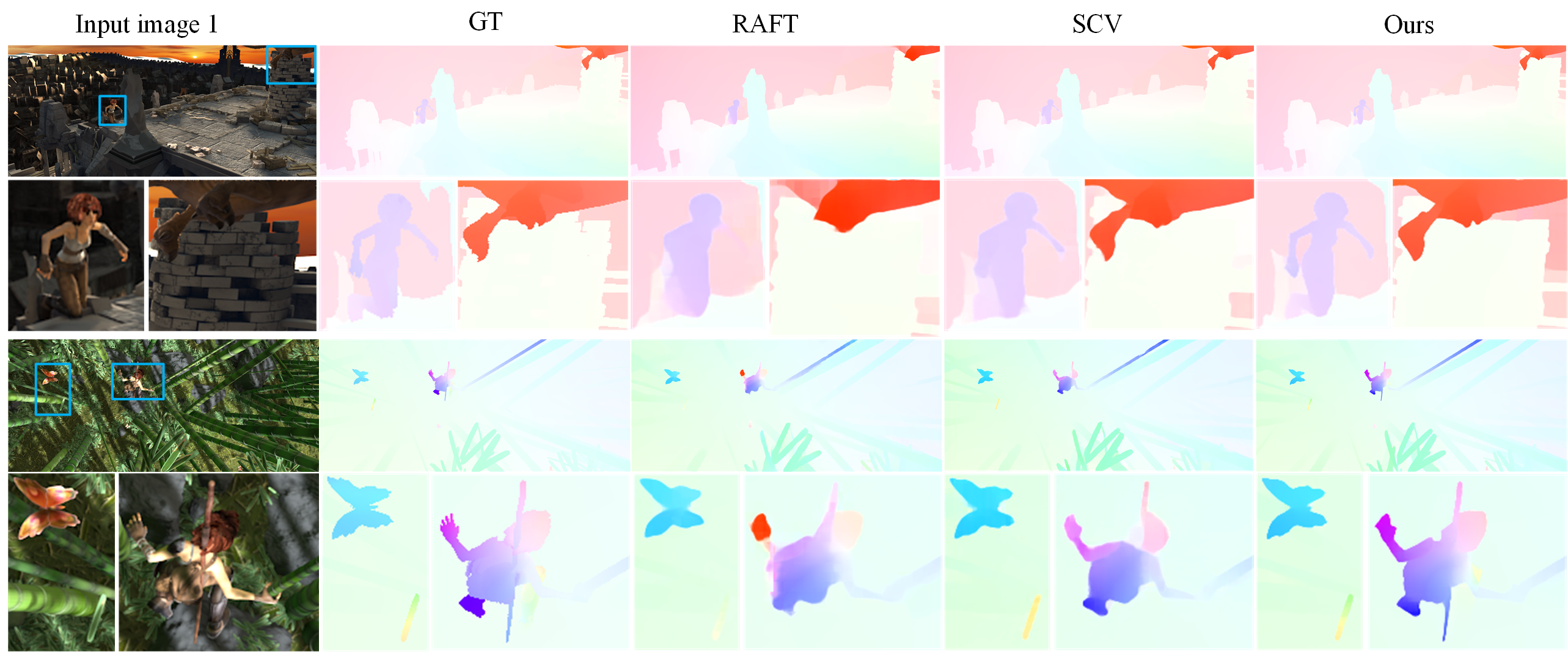} 
\caption{Visual comparison of optical flow estimates on the Sintel-Clean dataset. Compared with RAFT and SCV, our method performs particularly well, and our result is close to GT in the enlarged image frame. More results can be found in supplementary materials.}
\label{fig:sintel_res}
\end{figure*}
\subsection{Ablation study}
\label{sec:Ablation_study}
\paragraph{Correlation Volume} We first analyze the accuracy, memory and inference time of key components in our proposed method in \cref{tab:global_local}. In this comparative experiment, SCV(\textit{Sparse global})\cite{jiang2021learningSCV} is selected as a benchmark because it has low correlation volume in memory and state-of-the-art performance. In addition, we construct 4D correlation volumes with a resolution of (\textit{Dense global}) 1/16 and 1/4 resolution respectively, and each iteration performs a lookup like RAFT\cite{teed2020raft}. Using these benchmarks, we have conducted a partial experimental comparison. In the experiment, we implement a propagation experiment with a seed point of 4 and a local search experiment with a radius of 1 respectively. The results are clearly that only propagation(\textit{only p}) or local search(\textit{only ls}) has great advantages in terms of memory and speed at large resolutions, but the accuracy is reduced compared to the global method. The combination of propagation and local search (\textit{pm}) uses less time and memory to achieve comparable or better results than the global method. Especially, DIP consumes 10$\times$ less inference time than SCV on the size of 1088$\times$1920.
\vspace{-8pt}
\paragraph{Hyperparameters} Based on Patchmatch, we further experiment with hyperparameters and present them in \cref{tab:compare_pm_ls}. At first, the number of propagation seed points is set to 4, and the radius of local search is changed from 1 to 3. We can see that the accuracy is further improved when the search radius is increased from 1 to 2. When it is increased to 3, the accuracy is basically the same as radius 2, but the model inference time increases by $21\%$ . So the radius of the local search is fixed at 2. Then we change the number of propagation seed points from 4 to 8. However, the result is not improved significantly, but the model consumption increases. So we set the number of seed points to 4 for further optimization.
\vspace{-10pt}
\paragraph{Patchmatch and Inverse Patchmatch}Finally, we verified the effectiveness of the inverse Patchmatch and showed it in \cref{tab:pm_ipm}. In this experiment, we replaced the calculation method of correlation from propagation to inverse propagation, and adopted the previous training and evaluation strategy. The experiment shows that inverse propagation can achieve almost the same results as propagation. With a size of 1088$\times$1920, the inference time of inverse Patchmatch is reduced by 24$\%$ compared to Patchmatch.

In summary, based on our Patchmatch framework, we can achieve better performance with lower memory, and use inverse Patchmatch instead of Patchmatch to achieve the same performance with faster inference speed.
\begin{table}[htbp]
  \centering
  \resizebox{\linewidth}{!}{ 
  \begin{tabular}{ccccc}
    \toprule
    \multirow{2}[4]{*}{Method} & \multicolumn{2}{c}{Sintel(Train)} & \multicolumn{2}{c}{KITTI-15(train)} \\
 \cmidrule(lr){2-3} \cmidrule(lr){4-5}          & Clean & Final &EPE & F1-all \\
    \midrule
    HD3\cite{yin2019hierarchical}   & 3.84  & 8.77  & 13.17 & 24.0   \\
    LiteFlowNet\cite{hui2018liteflownet} & 2.48  & 4.04  & 10.39 & 28.50 \\
    PWC-Net\cite{sun2018pwc} & 2.55  & 3.93  & 10.35 & 33.7  \\
    LiteFlowNet2\cite{hui2020lightweight} & 2.24  & 3.78  & 8.97  & 25.90  \\
    VCN\cite{yang2019volumetric}   & 2.21  & 3.68  & 8.36  & 25.10  \\
    MaskFlowNet\cite{zhao2020maskflownet} & 2.25  & 3.61  & -     & 23.10   \\
    FlowNet2\cite{ilg2017flownet} & 2.02  & 3.54  & 10.08 & 30     \\
    DICL\cite{wang2020displacement}  & 1.94  & 3.77  & 8.70   & 23.60   \\
    RAFT\cite{teed2020raft}  & 1.43  &\textbf{2.71}  & \underline{5.04}  & \underline{17.40}   \\
    Flow1D\cite{xu2021high} & 1.98  & 3.27  & 6.69  & 22.95  \\
    SCV\cite{jiang2021learningSCV}   & \textbf{1.29}  & 2.95  & 6.80   & 19.30  \\
    ours  &  \underline{1.30}     & \underline{2.82}      &\textbf{4.29}  &\textbf{13.73}  \\
    \bottomrule
    \end{tabular}
    }
  \caption{Results on Sintel and KITTI. EPE refers to the average endpoint error and F1-all refers
to the percentage of optical flow outliers over all pixels. The best results are marked with bold and the second best results are marked with underline. Missing entries '-' indicates that the
result is not reported in the compared paper.}
  \label{tab:compare_table}
  \vspace{-8pt}
\end{table}
\begin{table*}[htbp]
  \centering
    \begin{tabular}{cccccccccc}

    \toprule
    \multicolumn{2}{c}{\multirow{3}[5]{*}{Method}} & \multicolumn{6}{c}{Sintel (test)}             & \multicolumn{2}{c}{KITTI-15 (test)} \\
\cmidrule{3-10}    \multicolumn{2}{c}{} & \multicolumn{3}{c}{Clean} & \multicolumn{3}{c}{Final} & \multicolumn{2}{c}{F1-all} \\
\cmidrule(lr){3-5}  \cmidrule(lr){6-8} \cmidrule(lr){9-10}
\multicolumn{2}{c}{} & EPE   & d0-10 & d10-60 & EPE   & d0-10 & d10-60 & All pixels & \multicolumn{1}{l}{Non-Occ pixels} \\
    \midrule
    \multirow{12}[0]{*}{2-view}     
    &FlowNet2\cite{ilg2017flownet} & 4.16  & 3.27  & 1.46  & 5.74  & 4.81  & 2.55  & 11.48  & 6.94 \\
    &PWC-Net+\cite{sun2019models} & 3.45  & 3.91  & 1.24  & 4.6   & 4.78  & 2.04  & 7.72 & 4.91 \\
    &LiteFlowNet2\cite{hui2020lightweight} & 3.48  & 3.27  & 1.43  & 4.69  & 4.04  & 1.89  & 7.74 & 4.42 \\
    &HD3\cite{yin2019hierarchical}   & 4.79  & 3.22  & 1.37  & 4.67  & 3.58  & 1.76  & 6.55 & - \\
    &VCN\cite{yang2019volumetric}   & 2.81  & 3.26  & 0.86  & 4.4   & 4.38  & 1.78  & 6.3 & 3.89 \\
    &MaskFlowNet\cite{zhao2020maskflownet} & 2.52  & 2.74  & 0.9   & 4.17  & 3.78  & 1.74  & 6.1 & 3.92 \\
    &ScopeFlow\cite{bar2020scopeflow} & 3.59  & 3.45  & 1.26  & 4.1   & 4.02  & 1.68  & 6.82 & 4.45 \\
    &DICL\cite{wang2020displacement}  & 2.12  & 2.2   & 0.58  & 3.44  & 3.27  & \underline{1.28}  & 6.31 & - \\
    &RAFT\cite{teed2020raft} & 1.94  &   -   &   -   & \textbf{3.18}  &   -   &   -   & \underline{5.1} & \underline{3.07} \\
    &Flow1D\cite{xu2021high} & 2.24  & 2.18   &0.87   & 3.81  &3.60    &1.75   & 6.27 & - \\
    &SCV\cite{jiang2021learningSCV} & \underline{1.72}  & \underline{1.39}  & \textbf{0.45}  & 3.6   & \underline{3.24}  & 1.42  & 6.17 & 3.43 \\
    &Ours & \textbf{1.67}  & \textbf{1.18}  & \textbf{0.45}  & \underline{3.22}   & \textbf{2.68}  & \textbf{1.23}  &\textbf{4.21}  &\textbf{2.43} \\
    \midrule 
    \multirow{3}[0]{*}{warm-start}  
    &RAFT & 1.61  & 1.62  & 0.51  & 2.86  & 3.11  & 1.13  & - & - \\
    &SCV & 1.77  &  -    &   -  & 3.88  &   -  &   -  & - & - \\
    &Ours & \textbf{1.44}  &\textbf{1.10}  &\textbf{0.41}  &\textbf{2.83}  &\textbf{2.72}  &\textbf{1.09}  & - & - \\
    \bottomrule
    \end{tabular}%
  \caption{Benchmark performance on Sintel and KITTI Test datasets. Missing entries '-' indicates that the
result is not reported in the compared paper and could not found on online benchmark. The best results are marked with bold and the second best results are marked with underline.}
  \label{tab:benchark}%
  \vspace{-8pt}
\end{table*}%

\subsection{Comparison with Existing Methods}
\label{sec:Comparison_with_Existing_Methods}
To demonstrate the superiority of our method, we have made a comprehensive comparison with the existing methods, including generalization, memory and special results.
\paragraph{Generalization}
\vspace{-12pt}
In order to verify the generalization of the model, we choose to use FlyingChairs\cite{dosovitskiy2015flownet} and FlyingThings3D\cite{mayer2016large} for training and Sintel\cite{butler2012naturalistic}, KITTI\cite{geiger2013vision} for test. Details are described in \Cref{sec:Training_schedule} and results are show in \cref{tab:compare_table}. Experiments show that our method exhibits strong generalization and achieves state-of-the-art results in the KITTI-15 dataset. Among them, F1-all is 13.73$\%$, reducing 21$\%$ from the best published result (17.4$\%$). On the Sintel dataset, we have also achieved results comparable to the state-of-the-art methods.
\vspace{-8pt}
\paragraph{Memory and High-resolution Results}
\vspace{-6pt}
\begin{figure}[htbp]
\centering 
\includegraphics[width=6cm]{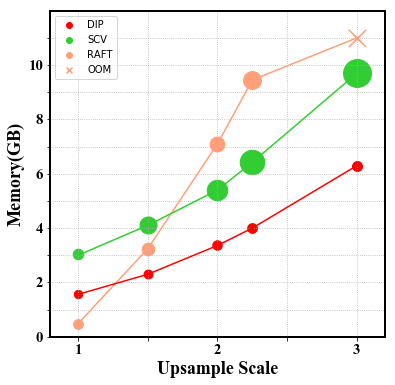} 
\caption{Upsampling to high-resolution size results. The memory limit is 11GB and the area of the bubbles is a mapping of the F1-all metric. We use upsampling of the KITTI dataset to evaluate memory and accuracy, and the resolution at the scale of 1 is 375 x 1242. 'OOM' means out of memory.}
\label{fig:memory}
\vspace{-8pt}
\end{figure}

We measure the accuracy and memory of different correlation volume algorithms at different resolutions in \cref{fig:memory}. Since there are few real and high-resolution datasets for the flow task, in the experiment we use the up-sampled kitti dataset for memory and accuracy evaluation. It can be seen that under the limitation of 11GB memory, the maximum output image scale of RAFT\cite{teed2020raft} is only 2.25. Moreover, the accuracy of SCV\cite{jiang2021learningSCV} is rapidly decreasing as the image scale increases. This demonstrates the effectiveness of our approach in saving memory and stabling accuracy when scaling correlation volumes to higher resolutions.

\paragraph{Benchmark Results}
\vspace{-8pt}
The performances of our DIP on the Sintel and KITTI-15 benchmarks are shown in \cref{tab:benchark}. We have achieved state-of-the-art results (1.72 $\rightarrow$ 1.67) on the Sintel-Clean dataset in the two-view case. Similar to RAFT, we also adopt the “warm-start” strategy which initialises current optical flow estimation with the flow estimates of the previous frame. On the Sintel-Clean benchmark our method ranks second for EPE. Compared with RAFT, we have improved the EPE from 1.61 to 1.44 (10.5$\%$ improvement). What's interesting is that our method achieves the best results on the \emph{d}$_{0-10}$ and \emph{d}$_{10-60}$, which shows that our method has obvious advantages in estimating the optical flow in small motion areas. \cref{fig:sintel_res} shows qualitative results of DIP on Sintel. Compared with RAFT and SCV, our results are much closer to the ground truth in the fine structure area.

On the KITTI-15 benchmark, our method ranks first on all the metrics among the published optical flow methods. Compared with RAFT, we have improved the F1-all from 3.07$\%$ to 2.43$\%$ (20.8$\%$ improvement) on Non-occluded pixels and the F1-all from 5.10$\%$ to 4.21$\%$ (17.5$\%$ improvement) on all pixels.

\section{Conclusion}
We propose a deep inverse Patchmatch framework for optical flow that focuses on reducing the computational cost and memory consumption of dense correlation volume. By reducing the computational and memory overhead, our model can work at a high-resolution and preserve the details of fine-structure. We also show a good trade-off between performance and cost. At the same time, we achieve comparable results with the state-of-the-art methods on public benchmarks and good generalization on different datasets. We believe that our inverse Patchmatch scheme can be used in more tasks, such as stereo matching, multi-view stereo vision and so on. In the future, more attention will be paid on the motion blur, large occlusion and other extreme scenes.

\clearpage
\bibliographystyle{ieee_fullname}
\bibliography{cvpr2022submission}

\begin{thebibliography}{10}\itemsep=-1pt

\bibitem{almatrafi2019davis}
Mohammed Almatrafi and Keigo Hirakawa.
\newblock Davis camera optical flow.
\newblock {\em IEEE Transactions on Computational Imaging}, 6:396--407, 2019.

\bibitem{bailer2017cnn}
Christian Bailer, Kiran Varanasi, and Didier Stricker.
\newblock Cnn-based patch matching for optical flow with thresholded hinge
  embedding loss.
\newblock In {\em Proceedings of the IEEE Conference on Computer Vision and
  Pattern Recognition}, pages 3250--3259, 2017.

\bibitem{bao2014fast}
Linchao Bao, Qingxiong Yang, and Hailin Jin.
\newblock Fast edge-preserving patchmatch for large displacement optical flow.
\newblock In {\em Proceedings of the IEEE Conference on Computer Vision and
  Pattern Recognition}, pages 3534--3541, 2014.

\bibitem{bar2020scopeflow}
Aviram Bar-Haim and Lior Wolf.
\newblock Scopeflow: Dynamic scene scoping for optical flow.
\newblock In {\em Proceedings of the IEEE/CVF Conference on Computer Vision and
  Pattern Recognition}, pages 7998--8007, 2020.

\bibitem{barnes2009Patchmatch}
Connelly Barnes, Eli Shechtman, Adam Finkelstein, and Dan~B Goldman.
\newblock Patchmatch: A randomized correspondence algorithm for structural
  image editing.
\newblock {\em ACM Trans. Graph.}, 28(3):24, 2009.

\bibitem{bleyer2011patchmatch}
Michael Bleyer, Christoph Rhemann, and Carsten Rother.
\newblock Patchmatch stereo-stereo matching with slanted support windows.
\newblock In {\em Bmvc}, volume~11, pages 1--11, 2011.

\bibitem{butler2012naturalistic}
Daniel~J Butler, Jonas Wulff, Garrett~B Stanley, and Michael~J Black.
\newblock A naturalistic open source movie for optical flow evaluation.
\newblock In {\em European conference on computer vision}, pages 611--625.
  Springer, 2012.

\bibitem{chang2013topology}
Jason Chang and John~W Fisher.
\newblock Topology-constrained layered tracking with latent flow.
\newblock In {\em Proceedings of the IEEE International Conference on Computer
  Vision}, pages 161--168, 2013.

\bibitem{chang2018pyramid}
Jia-Ren Chang and Yong-Sheng Chen.
\newblock Pyramid stereo matching network.
\newblock In {\em Proceedings of the IEEE Conference on Computer Vision and
  Pattern Recognition}, pages 5410--5418, 2018.

\bibitem{cheng2017segflow}
Jingchun Cheng, Yi-Hsuan Tsai, Shengjin Wang, and Ming-Hsuan Yang.
\newblock Segflow: Joint learning for video object segmentation and optical
  flow.
\newblock In {\em Proceedings of the IEEE international conference on computer
  vision}, pages 686--695, 2017.

\bibitem{dosovitskiy2015flownet}
Alexey Dosovitskiy, Philipp Fischer, Eddy Ilg, Philip Hausser, Caner Hazirbas,
  Vladimir Golkov, Patrick Van Der~Smagt, Daniel Cremers, and Thomas Brox.
\newblock Flownet: Learning optical flow with convolutional networks.
\newblock In {\em Proceedings of the IEEE international conference on computer
  vision}, pages 2758--2766, 2015.

\bibitem{duggal2019deeppruner}
Shivam Duggal, Shenlong Wang, Wei-Chiu Ma, Rui Hu, and Raquel Urtasun.
\newblock Deeppruner: Learning efficient stereo matching via differentiable
  patchmatch.
\newblock In {\em Proceedings of the IEEE/CVF International Conference on
  Computer Vision}, pages 4384--4393, 2019.

\bibitem{geiger2012we}
Andreas Geiger, Philip Lenz, and Raquel Urtasun.
\newblock Are we ready for autonomous driving? the kitti vision benchmark
  suite.
\newblock In {\em 2012 IEEE conference on computer vision and pattern
  recognition}, pages 3354--3361. IEEE, 2012.

\bibitem{guo2019group}
Xiaoyang Guo, Kai Yang, Wukui Yang, Xiaogang Wang, and Hongsheng Li.
\newblock Group-wise correlation stereo network.
\newblock In {\em Proceedings of the IEEE/CVF Conference on Computer Vision and
  Pattern Recognition}, pages 3273--3282, 2019.

\bibitem{hirschmuller2007stereo}
Heiko Hirschmuller.
\newblock Stereo processing by semiglobal matching and mutual information.
\newblock {\em IEEE Transactions on pattern analysis and machine intelligence},
  30(2):328--341, 2007.

\bibitem{ho2015optical}
HW Ho, Christophe De~Wagter, BDW Remes, and Guido~CHE de Croon.
\newblock Optical flow for self-supervised learning of obstacle appearance.
\newblock In {\em 2015 IEEE/RSJ International Conference on Intelligent Robots
  and Systems (IROS)}, pages 3098--3104. IEEE, 2015.

\bibitem{horn1981determining}
Berthold~KP Horn and Brian~G Schunck.
\newblock Determining optical flow.
\newblock {\em Artificial intelligence}, 17(1-3):185--203, 1981.

\bibitem{hosni2012fast}
Asmaa Hosni, Christoph Rhemann, Michael Bleyer, Carsten Rother, and Margrit
  Gelautz.
\newblock Fast cost-volume filtering for visual correspondence and beyond.
\newblock {\em IEEE Transactions on Pattern Analysis and Machine Intelligence},
  35(2):504--511, 2012.

\bibitem{hu2016efficient}
Yinlin Hu, Rui Song, and Yunsong Li.
\newblock Efficient coarse-to-fine patchmatch for large displacement optical
  flow.
\newblock In {\em Proceedings of the IEEE Conference on Computer Vision and
  Pattern Recognition}, pages 5704--5712, 2016.

\bibitem{hui2020liteflownet3}
Tak-Wai Hui and Chen~Change Loy.
\newblock Liteflownet3: Resolving correspondence ambiguity for more accurate
  optical flow estimation.
\newblock In {\em European Conference on Computer Vision}, pages 169--184.
  Springer, 2020.

\bibitem{hui2018liteflownet}
Tak-Wai Hui, Xiaoou Tang, and Chen~Change Loy.
\newblock Liteflownet: A lightweight convolutional neural network for optical
  flow estimation.
\newblock In {\em Proceedings of the IEEE conference on computer vision and
  pattern recognition}, pages 8981--8989, 2018.

\bibitem{hui2020lightweight}
Tak-Wai Hui, Xiaoou Tang, and Chen~Change Loy.
\newblock A lightweight optical flow cnn—revisiting data fidelity and
  regularization.
\newblock {\em IEEE transactions on pattern analysis and machine intelligence},
  43(8):2555--2569, 2020.

\bibitem{ilg2017flownet}
Eddy Ilg, Nikolaus Mayer, Tonmoy Saikia, Margret Keuper, Alexey Dosovitskiy,
  and Thomas Brox.
\newblock Flownet 2.0: Evolution of optical flow estimation with deep networks.
\newblock In {\em Proceedings of the IEEE conference on computer vision and
  pattern recognition}, pages 2462--2470, 2017.

\bibitem{jiang2021learning}
Shihao Jiang, Dylan Campbell, Yao Lu, Hongdong Li, and Richard Hartley.
\newblock Learning to estimate hidden motions with global motion aggregation.
\newblock {\em arXiv preprint arXiv:2104.02409}, 2021.

\bibitem{jiang2021learningSCV}
Shihao Jiang, Yao Lu, Hongdong Li, and Richard Hartley.
\newblock Learning optical flow from a few matches.
\newblock In {\em Proceedings of the IEEE/CVF Conference on Computer Vision and
  Pattern Recognition}, pages 16592--16600, 2021.

\bibitem{kondermann2016hci}
Daniel Kondermann, Rahul Nair, Katrin Honauer, Karsten Krispin, Jonas Andrulis,
  Alexander Brock, Burkhard Gussefeld, Mohsen Rahimimoghaddam, Sabine Hofmann,
  Claus Brenner, et~al.
\newblock The hci benchmark suite: Stereo and flow ground truth with
  uncertainties for urban autonomous driving.
\newblock In {\em Proceedings of the IEEE Conference on Computer Vision and
  Pattern Recognition Workshops}, pages 19--28, 2016.

\bibitem{kroeger2016fast}
Till Kroeger, Radu Timofte, Dengxin Dai, and Luc Van~Gool.
\newblock Fast optical flow using dense inverse search.
\newblock In {\em European Conference on Computer Vision}, pages 471--488.
  Springer, 2016.

\bibitem{kuang2017patchmatch}
Fangjun Kuang.
\newblock Patchmatch algorithms for motion estimation and stereo
  reconstruction.
\newblock Master's thesis, 2017.

\bibitem{loshchilov2017decoupled}
Ilya Loshchilov and Frank Hutter.
\newblock Decoupled weight decay regularization.
\newblock {\em arXiv preprint arXiv:1711.05101}, 2017.

\bibitem{lucas1981iterative}
Bruce~D Lucas, Takeo Kanade, and Others.
\newblock {An iterative image registration technique with an application to
  stereo vision}.
\newblock In {\em Proc. of the Intl. Joint Conference on Artificial
  Intelligence}, volume~81, pages 674--679, 1981.

\bibitem{makansi2017end}
Osama Makansi, Eddy Ilg, and Thomas Brox.
\newblock End-to-end learning of video super-resolution with motion
  compensation.
\newblock In {\em German conference on pattern recognition}, pages 203--214.
  Springer, 2017.

\bibitem{mayer2016large}
Nikolaus Mayer, Eddy Ilg, Philip Hausser, Philipp Fischer, Daniel Cremers,
  Alexey Dosovitskiy, and Thomas Brox.
\newblock A large dataset to train convolutional networks for disparity,
  optical flow, and scene flow estimation.
\newblock In {\em Proceedings of the IEEE conference on computer vision and
  pattern recognition}, pages 4040--4048, 2016.

\bibitem{menze2015object}
Moritz Menze and Andreas Geiger.
\newblock Object scene flow for autonomous vehicles.
\newblock In {\em Proceedings of the IEEE conference on computer vision and
  pattern recognition}, pages 3061--3070, 2015.

\bibitem{geiger2013vision}
Moritz Menze, Christian Heipke, and Andreas Geiger.
\newblock Joint 3d estimation of vehicles and scene flow.
\newblock {\em ISPRS annals of the photogrammetry, remote sensing and spatial
  information sciences}, 2:427, 2015.

\bibitem{paszke2019pytorch}
Adam Paszke, Sam Gross, Francisco Massa, Adam Lerer, James Bradbury, Gregory
  Chanan, Trevor Killeen, Zeming Lin, Natalia Gimelshein, Luca Antiga, et~al.
\newblock Pytorch: An imperative style, high-performance deep learning library.
\newblock {\em Advances in neural information processing systems},
  32:8026--8037, 2019.

\bibitem{revaud2016deepmatching}
Jerome Revaud, Philippe Weinzaepfel, Zaid Harchaoui, and Cordelia Schmid.
\newblock Deepmatching: Hierarchical deformable dense matching.
\newblock {\em International Journal of Computer Vision}, 120(3):300--323,
  2016.

\bibitem{scharstein2014high}
Daniel Scharstein, Heiko Hirschm{\"u}ller, York Kitajima, Greg Krathwohl, Nera
  Ne{\v{s}}i{\'c}, Xi Wang, and Porter Westling.
\newblock High-resolution stereo datasets with subpixel-accurate ground truth.
\newblock In {\em German conference on pattern recognition}, pages 31--42.
  Springer, 2014.

\bibitem{schops2017multi}
Thomas Schops, Johannes~L Schonberger, Silvano Galliani, Torsten Sattler,
  Konrad Schindler, Marc Pollefeys, and Andreas Geiger.
\newblock A multi-view stereo benchmark with high-resolution images and
  multi-camera videos.
\newblock In {\em Proceedings of the IEEE Conference on Computer Vision and
  Pattern Recognition}, pages 3260--3269, 2017.

\bibitem{shen2021cfnet}
Zhelun Shen, Yuchao Dai, and Zhibo Rao.
\newblock Cfnet: Cascade and fused cost volume for robust stereo matching.
\newblock In {\em Proceedings of the IEEE/CVF Conference on Computer Vision and
  Pattern Recognition}, pages 13906--13915, 2021.

\bibitem{simonyan2014two}
Karen Simonyan and Andrew Zisserman.
\newblock Two-stream convolutional networks for action recognition in videos.
\newblock {\em arXiv preprint arXiv:1406.2199}, 2014.

\bibitem{smith2019super}
Leslie~N Smith and Nicholay Topin.
\newblock Super-convergence: Very fast training of neural networks using large
  learning rates.
\newblock In {\em Artificial Intelligence and Machine Learning for Multi-Domain
  Operations Applications}, volume 11006, page 1100612. International Society
  for Optics and Photonics, 2019.

\bibitem{sun2013fully}
Deqing Sun, Jonas Wulff, Erik~B Sudderth, Hanspeter Pfister, and Michael~J
  Black.
\newblock A fully-connected layered model of foreground and background flow.
\newblock In {\em Proceedings of the IEEE Conference on Computer Vision and
  Pattern Recognition}, pages 2451--2458, 2013.

\bibitem{sun2018pwc}
Deqing Sun, Xiaodong Yang, Ming-Yu Liu, and Jan Kautz.
\newblock Pwc-net: Cnns for optical flow using pyramid, warping, and cost
  volume.
\newblock In {\em Proceedings of the IEEE conference on computer vision and
  pattern recognition}, pages 8934--8943, 2018.

\bibitem{sun2019models}
Deqing Sun, Xiaodong Yang, Ming-Yu Liu, and Jan Kautz.
\newblock Models matter, so does training: An empirical study of cnns for
  optical flow estimation.
\newblock {\em IEEE transactions on pattern analysis and machine intelligence},
  42(6):1408--1423, 2019.

\bibitem{teed2020raft}
Zachary Teed and Jia Deng.
\newblock Raft: Recurrent all-pairs field transforms for optical flow.
\newblock In {\em European conference on computer vision}, pages 402--419.
  Springer, 2020.

\bibitem{tsai2016video}
Yi-Hsuan Tsai, Ming-Hsuan Yang, and Michael~J Black.
\newblock Video segmentation via object flow.
\newblock In {\em Proceedings of the IEEE conference on computer vision and
  pattern recognition}, pages 3899--3908, 2016.

\bibitem{tu2017fusing}
Zhigang Tu, Zuwei Guo, Wei Xie, Mengjia Yan, Remco~C Veltkamp, Baoxin Li, and
  Junsong Yuan.
\newblock Fusing disparate object signatures for salient object detection in
  video.
\newblock {\em Pattern Recognition}, 72:285--299, 2017.

\bibitem{tu2019survey}
Zhigang Tu, Wei Xie, Dejun Zhang, Ronald Poppe, Remco~C Veltkamp, Baoxin Li,
  and Junsong Yuan.
\newblock A survey of variational and cnn-based optical flow techniques.
\newblock {\em Signal Processing: Image Communication}, 72:9--24, 2019.

\bibitem{wang2021patchmatchnet}
Fangjinhua Wang, Silvano Galliani, Christoph Vogel, Pablo Speciale, and Marc
  Pollefeys.
\newblock Patchmatchnet: Learned multi-view patchmatch stereo.
\newblock In {\em Proceedings of the IEEE/CVF Conference on Computer Vision and
  Pattern Recognition}, pages 14194--14203, 2021.

\bibitem{wang2020displacement}
Jianyuan Wang, Yiran Zhong, Yuchao Dai, Kaihao Zhang, Pan Ji, and Hongdong Li.
\newblock Displacement-invariant matching cost learning for accurate optical
  flow estimation.
\newblock {\em arXiv preprint arXiv:2010.14851}, 2020.

\bibitem{weinzaepfel2013deepflow}
Philippe Weinzaepfel, Jerome Revaud, Zaid Harchaoui, and Cordelia Schmid.
\newblock Deepflow: Large displacement optical flow with deep matching.
\newblock In {\em Proceedings of the IEEE international conference on computer
  vision}, pages 1385--1392, 2013.

\bibitem{xiao2016track}
Fanyi Xiao and Yong Jae~Lee.
\newblock Track and segment: An iterative unsupervised approach for video
  object proposals.
\newblock In {\em Proceedings of the IEEE conference on computer vision and
  pattern recognition}, pages 933--942, 2016.

\bibitem{xu2021high}
Haofei Xu, Jiaolong Yang, Jianfei Cai, Juyong Zhang, and Xin Tong.
\newblock High-resolution optical flow from 1d attention and correlation.
\newblock In {\em Proceedings of the IEEE/CVF International Conference on
  Computer Vision}, pages 10498--10507, 2021.

\bibitem{yang2019volumetric}
Gengshan Yang and Deva Ramanan.
\newblock Volumetric correspondence networks for optical flow.
\newblock {\em Advances in neural information processing systems}, 32:794--805,
  2019.

\bibitem{yin2019hierarchical}
Zhichao Yin, Trevor Darrell, and Fisher Yu.
\newblock Hierarchical discrete distribution decomposition for match density
  estimation.
\newblock In {\em Proceedings of the IEEE/CVF Conference on Computer Vision and
  Pattern Recognition}, pages 6044--6053, 2019.

\bibitem{zhang2019ga}
Feihu Zhang, Victor Prisacariu, Ruigang Yang, and Philip~HS Torr.
\newblock Ga-net: Guided aggregation net for end-to-end stereo matching.
\newblock In {\em Proceedings of the IEEE/CVF Conference on Computer Vision and
  Pattern Recognition}, pages 185--194, 2019.

\bibitem{zhang2020domain}
Feihu Zhang, Xiaojuan Qi, Ruigang Yang, Victor Prisacariu, Benjamin Wah, and
  Philip Torr.
\newblock Domain-invariant stereo matching networks.
\newblock In {\em European Conference on Computer Vision}, pages 420--439.
  Springer, 2020.

\bibitem{zhao2020maskflownet}
Shengyu Zhao, Yilun Sheng, Yue Dong, Eric~I Chang, Yan Xu, et~al.
\newblock Maskflownet: Asymmetric feature matching with learnable occlusion
  mask.
\newblock In {\em Proceedings of the IEEE/CVF Conference on Computer Vision and
  Pattern Recognition}, pages 6278--6287, 2020.

\end{thebibliography}
\clearpage

\section*{Appendix}
\renewcommand{\thesection}{\Alph{section}}
\renewcommand{\thetable}{\Alph{table}}
\renewcommand{\thefigure}{\Alph{figure}}
\setcounter{section}{0}
\setcounter{table}{0}
\setcounter{figure}{0}

\section{Patchmatch in Flow}
\label{sec: patchmatch in flow}
The traditional Patchmatch methods\cite{barnes2009Patchmatch} consists of three components: Random Initialization, Propagation and Random Search. 

\begin{figure}[htbp]
\centering
\begin{subfigure}[t]{0.48\textwidth}
\centering
\includegraphics[width=\textwidth]{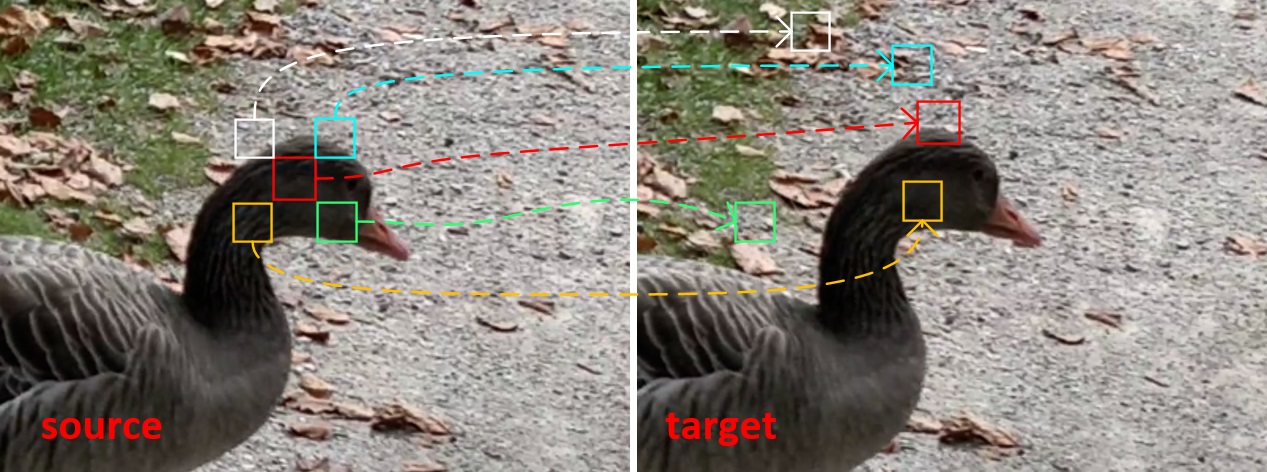}
\caption{Random Initialization}
\label{fig:pm_init}
\end{subfigure}
\begin{subfigure}[t]{0.48\textwidth}
\centering
\includegraphics[width=\textwidth]{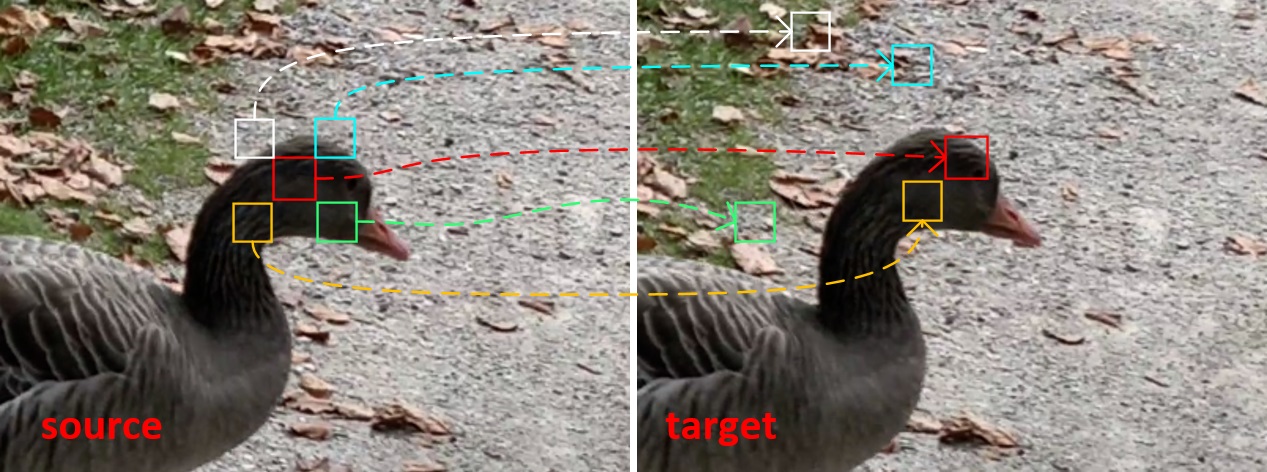}
\caption{Propagation}
\label{fig:pm_p}
\end{subfigure}
\begin{subfigure}[t]{0.48\textwidth}
\centering
\includegraphics[width=\textwidth]{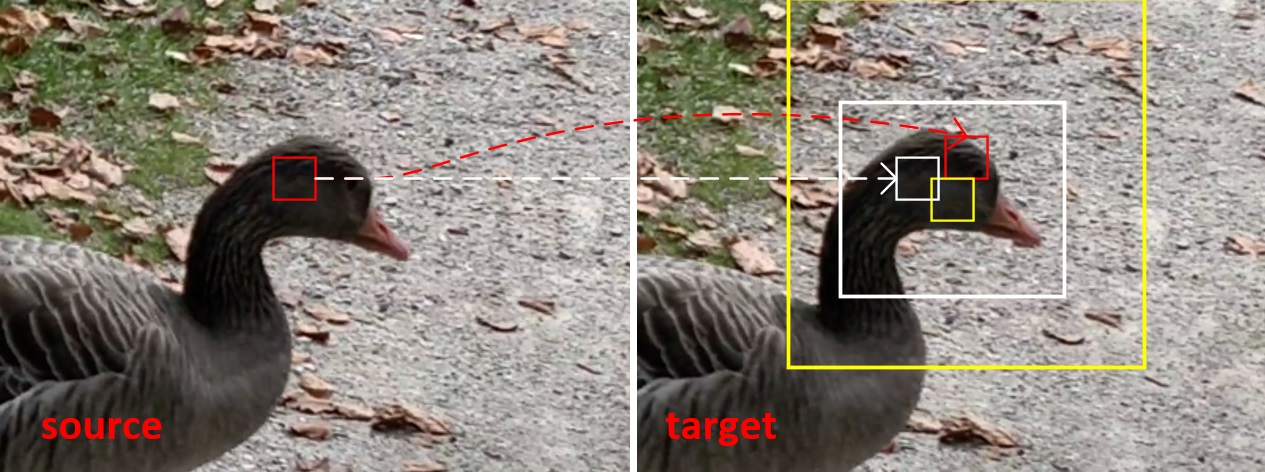}
\caption{Random Search}
\label{fig:pm_rs}
\end{subfigure}
\begin{subfigure}[t]{0.48\textwidth}
\centering
\includegraphics[width=\textwidth]{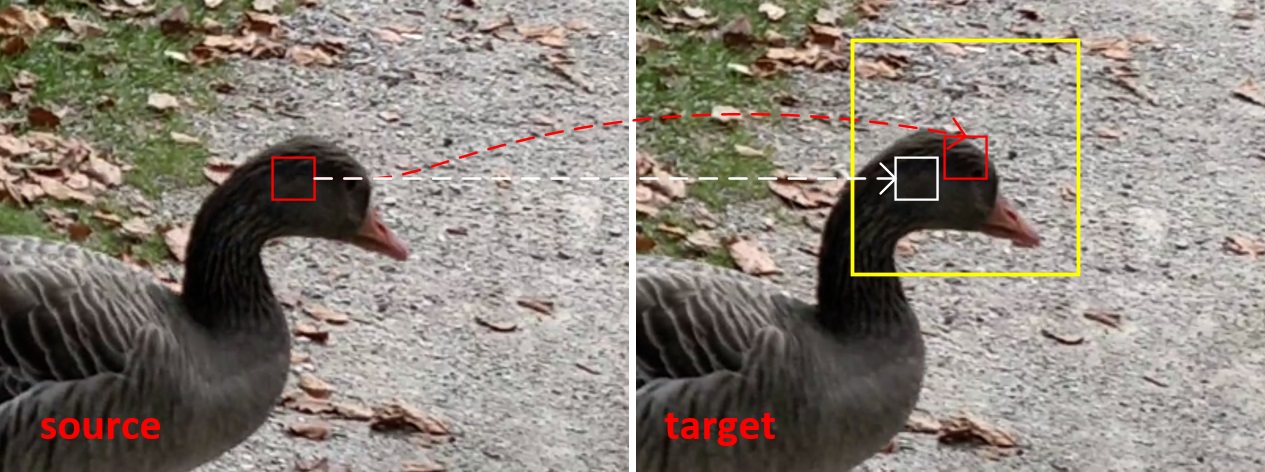}
\caption{Local Search}
\label{fig:pm_ls}
\end{subfigure}
\caption{A toy example for the Patchmatch in flow}
\label{fig:pmf}
\vspace{-3mm}
\end{figure}

In the initialization stage, the flow is initialized either randomly or based on some prior information. A toy example for this stage is shown in \cref{fig:pm_init}, the flow is initialized randomly. So for a patch represented by the red box with its 4 neighbors represented by the white, blue, yellow and green box respectively in the source image, the random flow relation can be represented as the dotted arrows to the target patches. That is to say, the red box in the source image moves to the red box in the target image with a random flow. In DIP, the flow is initialized randomly at the begining and after getting the flow at a 1/16 resolution, we use it as an initial flow at the 1/4 stage.

In the propagation stage, every patch compares the costs of its own flow with that of its neighbors and updates them if the flow of its neighbors lead to a lower cost. As the \cref{fig:pm_p} shows, after the initialization, for the red box, the flows from itself and its neighbors will be used to compute 5 correlation volume, and it is obvious that the flow candidate from the yellow box results in the maxmium correlation. So the flow of the red box will be update to the flow from the yellow box. In order to make the propagation stage friendly to the end-to-end pipeline, we shift the flow map toward the 4 neighbors(top-left, top-right, bottom-left, botton-right) so that we can use the flow from the 4 neighbors to compute the corresponding correlation by a vectorization operator. For example, when shifting the flow to the down-right, the point(1,1) will get the flow of point(0,0), the correlation at point(1,1) actually is computed by the flow at point(0,0). After shifting 4 times, we can get 5 correlation coefficients for point(1, 1) based on the flow from point(1, 1), (0,0), (0,2), (2,0), (2,2). Then we can choose the best flow for point(1, 1) according to correlation volume. 

The random search step is an essential step to make Patchmatch work. Propagation can converge very quickly but often end up in a local minimum. So it is necessary to introduce new information into the pipeline. In the random search stage, it is achieved by selecting a flow candidate randomly from an interval, whose length decreases exponentially with respect to the number of searches. Just like the \cref{fig:pm_p} shows, the flow of the red box is updated and is closer to the good match, but it is not the best match. So it is necessary to add the random search stage to get more flow candidates further. As the \cref{fig:pm_rs} shows, the candidates can be searched in the target image by a binary random search method. Centered on the red box, the first random search will be done within the big yellow box whose radius is min(imagewidth/2, imageheight/2), and the better match can be found at the small yellow box(if the small yellow box gets a worse match, the flow won't be updated). So the next random search will be done centered with the small yellow box within the big white box, and luckily the random search gets the small white box which is much better than the small yellow box and is extremely close to the best match. So after this stage, the flow for the red box is updated to the motion with the small white box which is represented by the white dotted arrows. However, random search is not friendy to the deep learning pipeline. So we replace this stage with a local search method, which aggregates the flow candidates from a 5x5 windows on the 1/16 resolution coarsely and the 1/4 resolution finely. It can be also represented by a toy example shown as the \cref{fig:pm_ls}, the good match can be found by aggregrating within the yellow box. And experiments also confirm that this alternative works well.

It is recommend to refer the work\cite{kuang2017patchmatch}, they make a good summary of Patchmatch and application to stereo task.

\section{Domain-invariance in Stereo Matching}
\label{sec:Stereo}
\begin{figure}[htbp]
\centering 
\includegraphics[width=8cm]{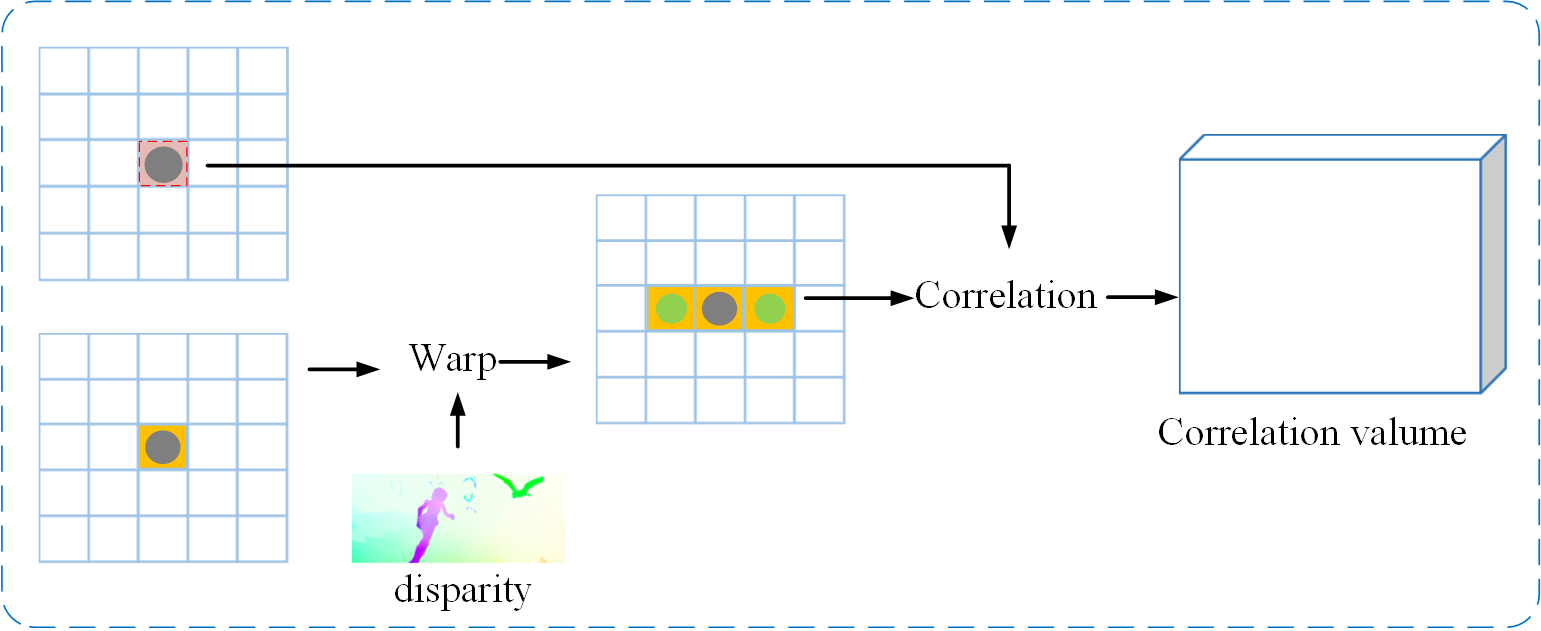} 
\caption{Local Search block for Stereo.}
\label{fig:LSS}
\vspace{-3mm}
\end{figure}

In this supplementary document, we first applied DIP to Stereo to demonstrate the portability. The core of the stereo matching algorithm is to obtain a dense disparity map of a pair of rectified stereo images, where disparity refers to the horizontal relationship between a pair of corresponding pixels on the left and right images. Optical flow and stereo are closely related problems. The difference is that optical flow predicts the displacement of the pixel in the plane, while stereo only needs to estimate the displacement of the pixel in a horizontal line. Therefore, we improved the local search block in DIP to make it more relevant to stereo task. Specifically, we reduced the search range of local search block from 2D search to 1D search. The entire local search block for Stereo is shown in \cref{fig:LSS}.

In the main paper we have proved that inverse patchmatch and local search in optical flow not only obtain high-precision results but also have strong domain-invariance. In the stereo matching experiments, we follow the training strategy of DSMNet\cite{zhang2020domain}, which is to train only on the Sceneflow dataset \cite{mayer2016large}, and other real datasets (such as Kitti\cite{geiger2012we, menze2015object}, Middlebury\cite{scharstein2014high}, and ETH3D\cite{schops2017multi}) are used to evaluate the cross-domain generalization ability of the network. Before training, the input images are randomly cropped to 384 × 768, and the pixel intensity is normalized to -1 and 1. We train the model on the Sceneflow dataset for 160K steps with a OneCycle learning rate schedule of initial learning rate is 0.0004.

\begin{table}[htbp]
  \centering
    \begin{tabular}{cccccc}
    \toprule
    \multirow{2}[4]{*}{Models} & \multicolumn{2}{c}{KITTI} & \multicolumn{2}{c}{Middlebury} & \multicolumn{1}{c}{\multirow{2}[4]{*}{ETH3D}} \\
\cmidrule{2-5}    \multicolumn{1}{c}{} & 2012  & 2015  & \multicolumn{1}{c}{half} & \multicolumn{1}{c}{quarter} &  \\
    \midrule
    CostFilter\cite{hosni2012fast} & 21.7  & 18.9  & 40.5  & 17.6  & 31.1 \\
    \midrule
    PatchMatch\cite{bleyer2011patchmatch}& 20.1  & 17.2  & 38.6  & 16.1  & 24.1 \\
    \midrule
    SGM\cite{hirschmuller2007stereo} & 7.1   & 7.6   & 25.2  & 10.7  & 12.9 \\
    \midrule
    Training set & \multicolumn{5}{c}{SceneFlow} \\
    \midrule
    HD3\cite{yin2019hierarchical} & 23.6  & 26.5  & 37.9  & 20.3  & 54.2 \\
    \midrule
    PSMNet\cite{chang2018pyramid} & 15.1  & 16.3  & 25.1  & 14.2  & 23.8 \\
    \midrule
    Gwcnet\cite{guo2019group} & 12.5  & 12.6  & 34.2  & 18.1 & 30.1 \\
    \midrule
    GANet\cite{zhang2019ga} & 10.1  & 11.7  & 20.3  & 11.2  & 14.1 \\
    \midrule
    DSMNet\cite{zhang2020domain} & 6.2   & 6.5   & \textbf{13.8}  & \textbf{8.1}   & 6.2 \\
    \midrule
    CFNet\cite{shen2021cfnet} & \textbf{4.7}   & 5.8  & 21.2  & 13.1   & 5.8 \\    
    \midrule
    \textbf{Ours-Flow} & 5.6 & \underline{5.7} & 17.2 & 10.6 & \underline{5.5} \\
    \textbf{Ours-Stereo} & \underline{4.9} & \textbf{4.9} & \underline{14.9} & \underline{8.8} & \textbf{3.3} \\
   
    \bottomrule
    \end{tabular}%
  
    \caption{Comparing with other advanced methods on KITTI, Middlebury and ETH3D training sets. All methods were trained on SceneFlow. Errors are the percent of pixels with end-point-error greater than the specified threshold. We use the standard evaluation thresholds: 3px for KITTI, 2px for Middlebury, 1px for ETH3D.
    \label{tab:compare}%
}
\end{table}%

\vspace{-10pt}
\paragraph{Domain-invariance ability} The domain-invariance is an ability that generalizes to unseen data without training. In \cref{tab:compare}, we compare our DIP with other state-of-the-art deep neural network models on the four unseen real-world datasets. All the models are trained on SceneFlow data. On the KITTI and ETH3D dataset our result  far outperforms the previous methods. In the Middlebury dataset, our results only lag behind DSMNet better than all the other methods. Compared to DIP-Flow, DIP-Stereo has more domain-invariance capability, which indicates that our proposed local search block for Stereo is effective in handling Stereo tasks.

\begin{figure*}[htbp]
 \begin{minipage}{0.247\linewidth}
     \centerline{\includegraphics[width=\textwidth]{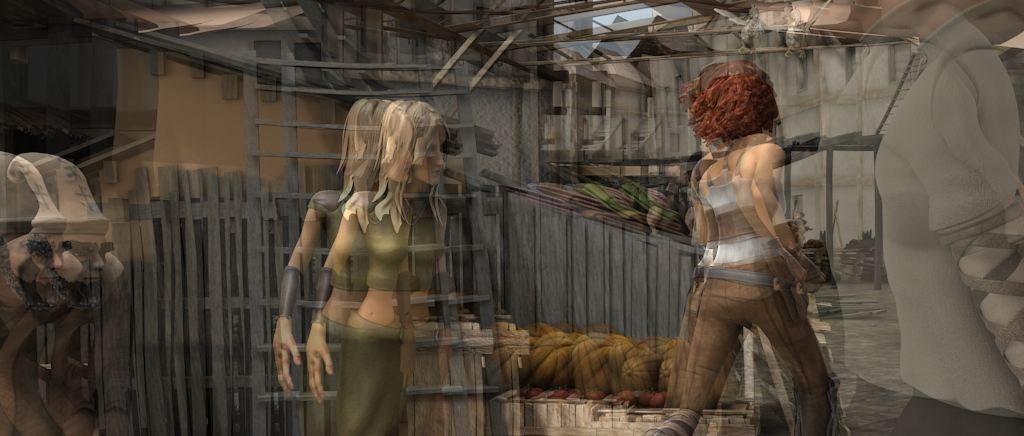}}
     \centerline{\includegraphics[width=\textwidth]{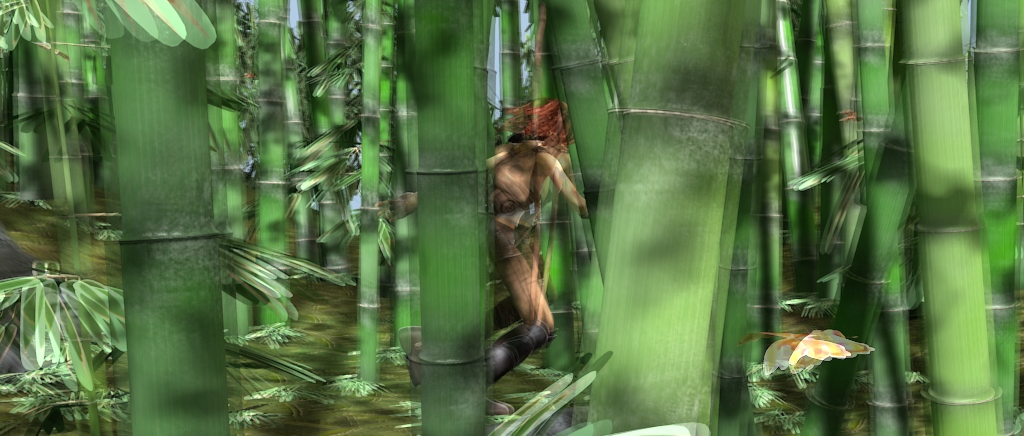}}
     \centerline{\includegraphics[width=\textwidth]{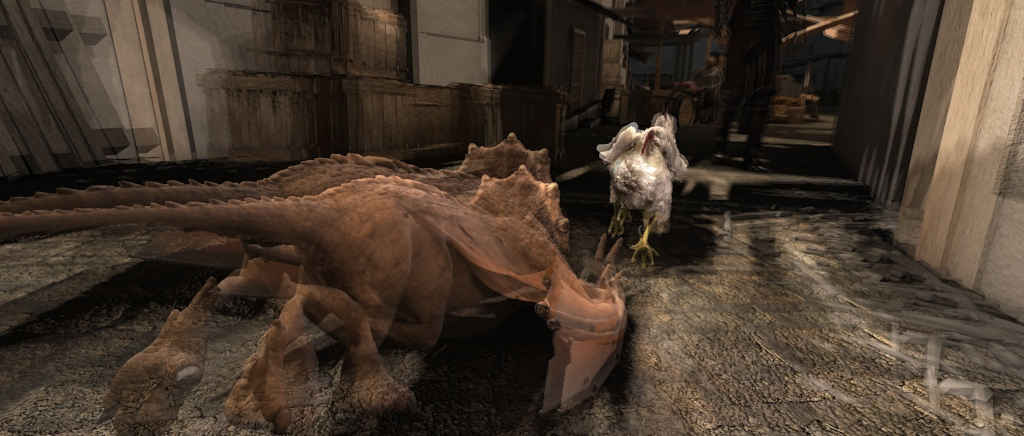}}
     \centerline{Image Overlay}
 \end{minipage}
 \begin{minipage}{0.247\linewidth}
     \centerline{\includegraphics[width=\textwidth]{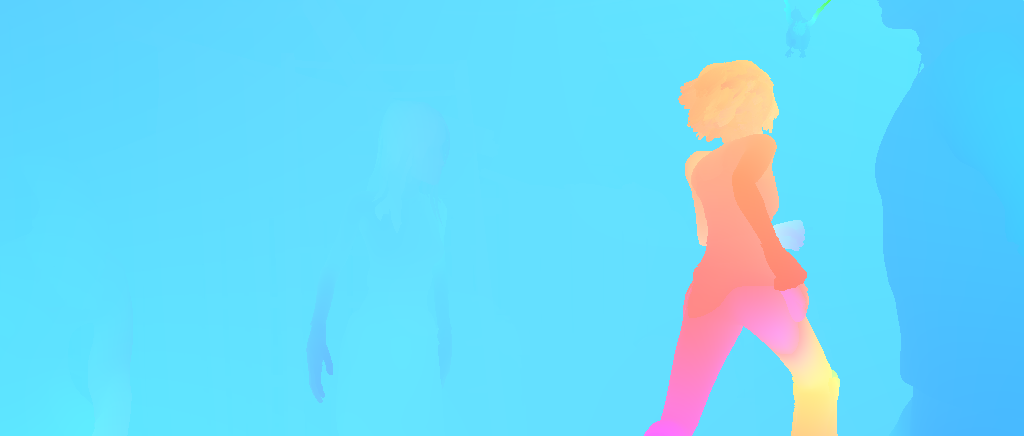}}
     \centerline{\includegraphics[width=\textwidth]{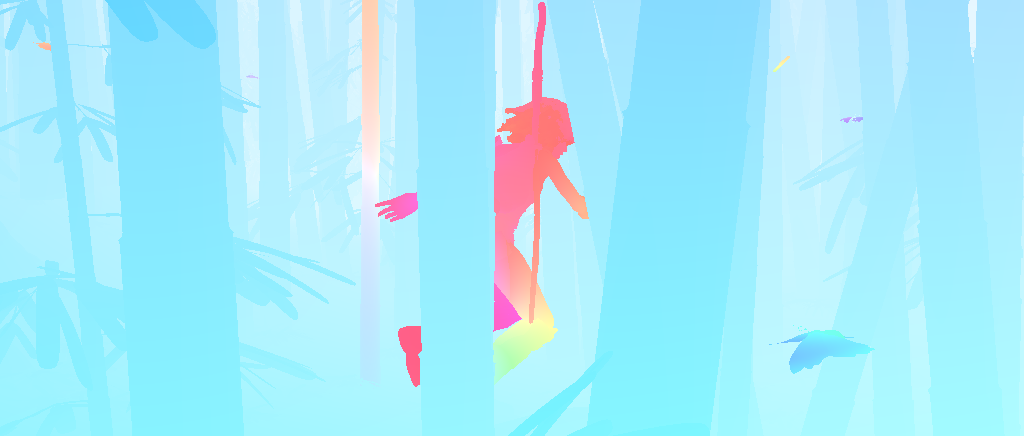}}
     \centerline{\includegraphics[width=\textwidth]{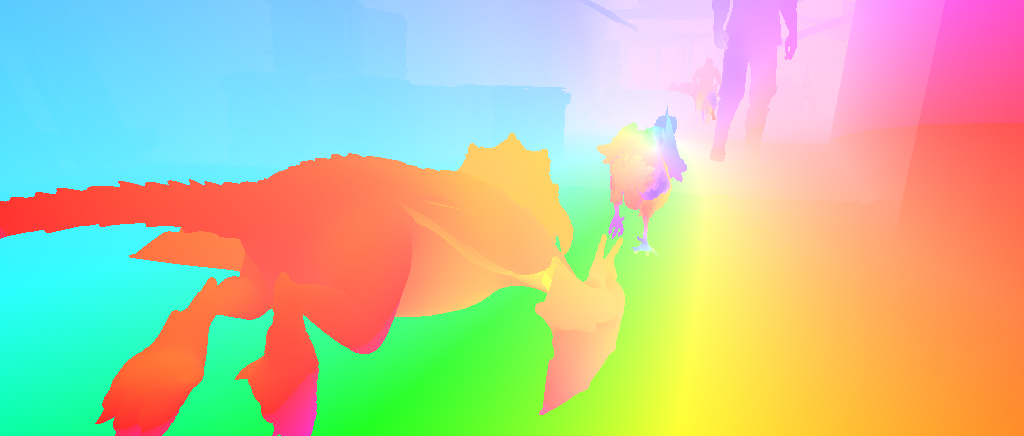}}
     \centerline{Ground truth}
 \end{minipage}
 \begin{minipage}{0.247\linewidth}
     \centerline{\includegraphics[width=\textwidth]{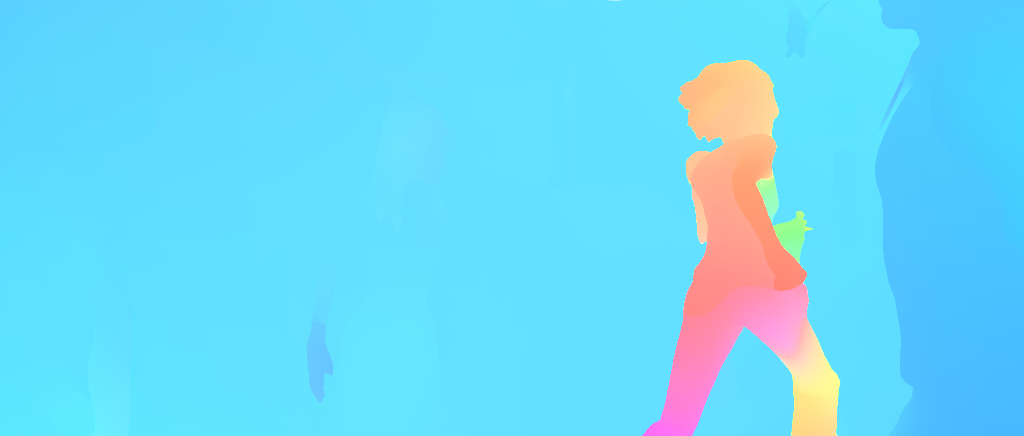}}
     \centerline{\includegraphics[width=\textwidth]{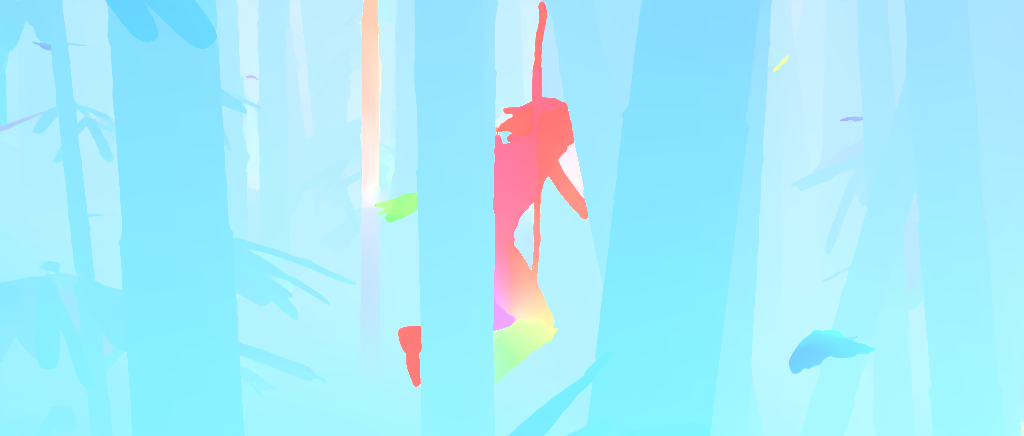}}
     \centerline{\includegraphics[width=\textwidth]{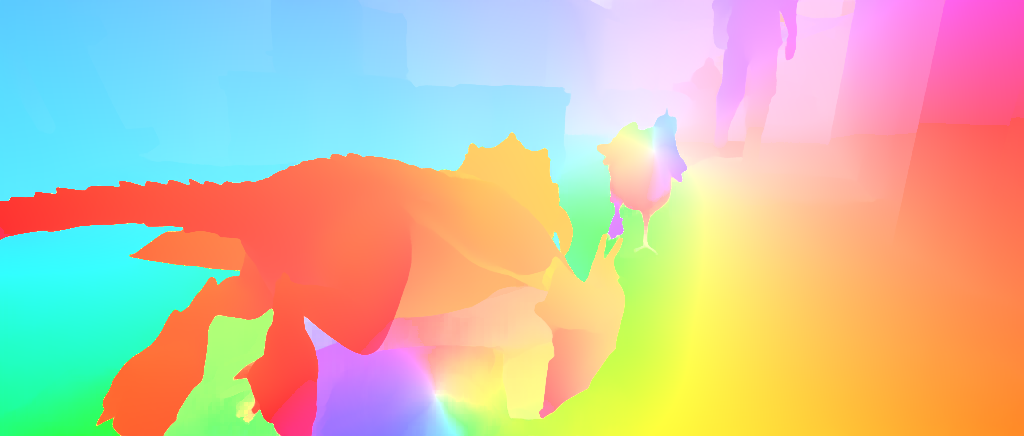}}
     \centerline{Two Layers}
 \end{minipage}
 \begin{minipage}{0.247\linewidth}
     \centerline{\includegraphics[width=\textwidth]{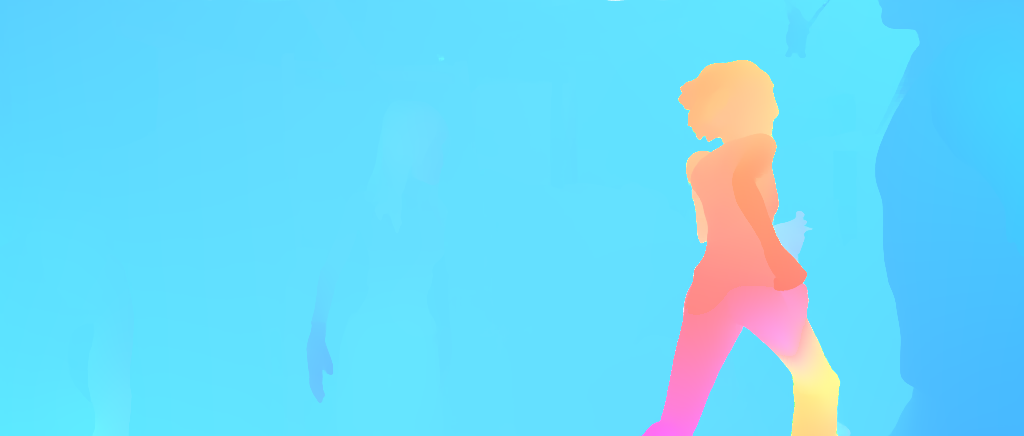}}
     \centerline{\includegraphics[width=\textwidth]{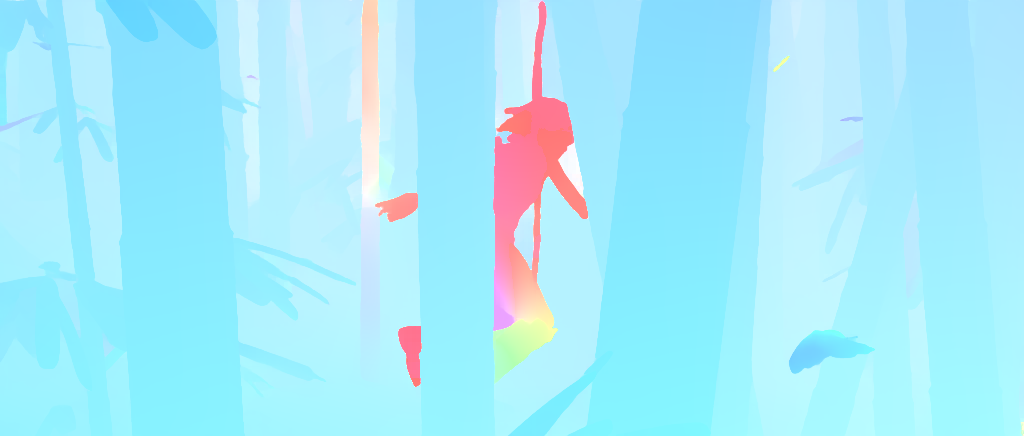}}
     \centerline{\includegraphics[width=\textwidth]{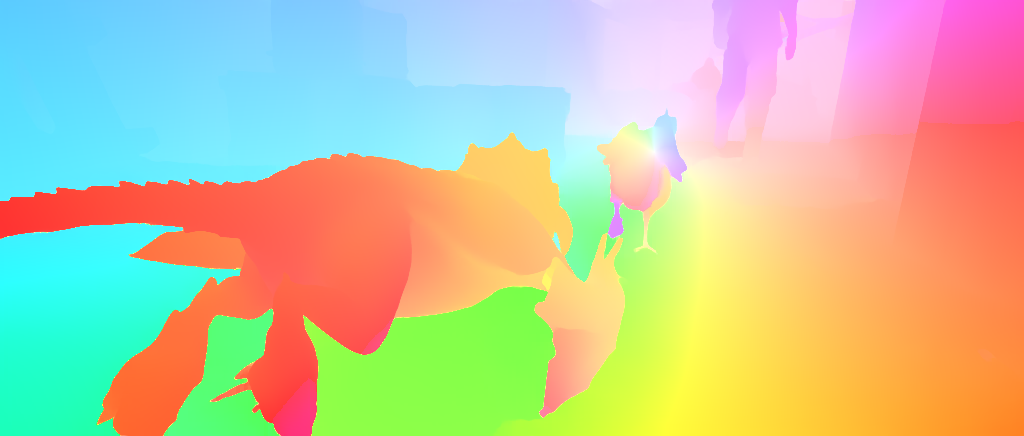}}
     \centerline{Adaptive Layers}
 \end{minipage}

\caption{Results compare between fixed two layers and adaptive layers. The two-level pyramid adopts a strategy from 1/16 to 1/4 resolution. The adaptive way adaptively selects the initial resolution according to the initial optical flow, such as 1/16, 1/8, or 1/4 initial resolution.}
\label{fig:adpc}
\end{figure*}

\begin{figure*}[htbp]
\centering 
\includegraphics[width=17cm]{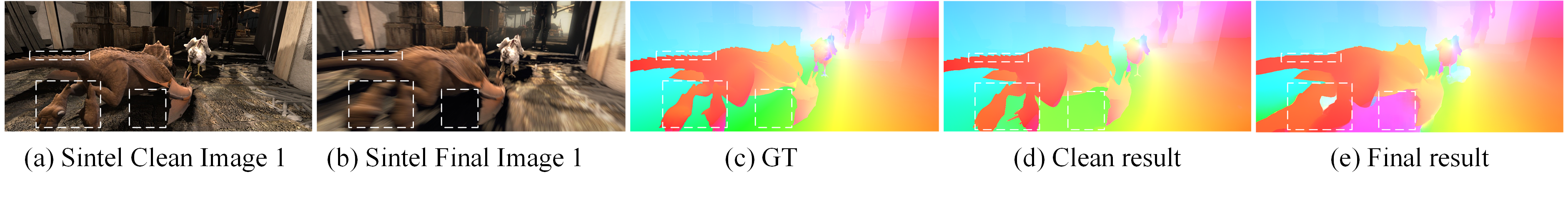} 
\caption{Comparison of results between normal scenes and motion blur scenes. Motion blur causes incorrect optical flow estimation.}
\label{fig:limit}
\end{figure*}

\section{Adaptive Layers}
\label{sec: Limitations}
Because DIP uses the same process and parameters for each pyramid, we can define any pyramid layers to make predictions, instead of using only two layers pyramid as we trained. Experiments show that when multilayer pyramid prediction is used, a more accurate optical flow can be obtained. Especially for continuous optical flow prediction, the adaptive pyramid layers can be used to obtain better results.

DIP supports initializing optical flow input. In the optical flow prediction of consecutive frames of video, we can take the forward interpolation of the previous result as the initialization input of the current frame. If the maximum displacement of the initialized optical flow is large, the motion of the current frame may also be large, at which point we need to start from a low-resolution layer. And to ensure accuracy, the sampling rate of the pyramid is 2 instead of 4. If previous displacement is very small, the motion of the current frame may also be small, at which point we need only one layer of pyramid prediction. \cref{fig:adpc} shows the comparison between the two-layers pyramid and the adaptive layers pyramid, and both initialize using the “warm-start” strategy.

\section{More Results on High-Resolution}
\label{sec:High-res}
\begin{figure*}[htbp]
 \begin{minipage}{0.498\linewidth}
     \centerline{\includegraphics[width=\textwidth]{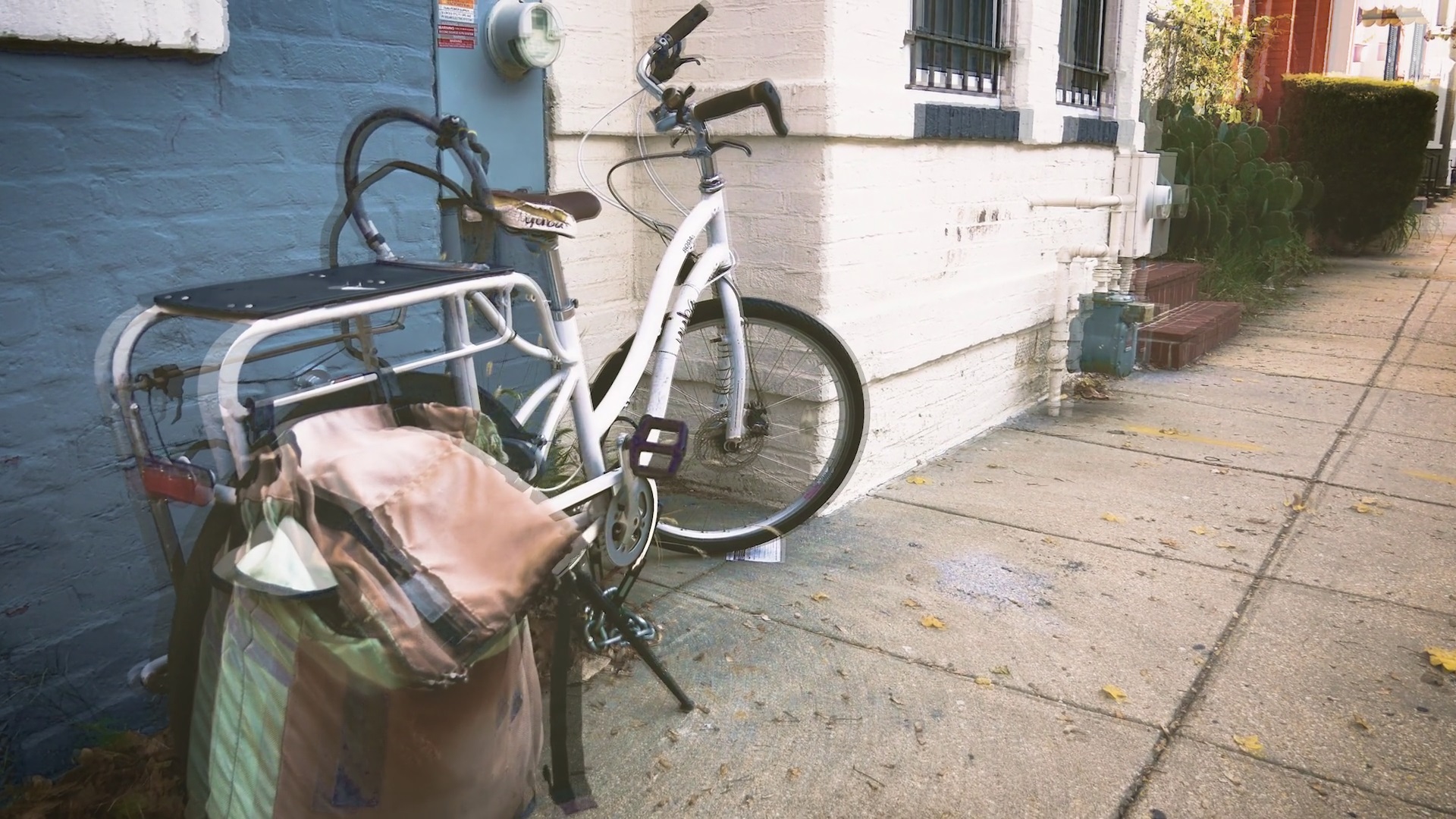}}
     \vspace{1pt}
     \centerline{\includegraphics[width=\textwidth]{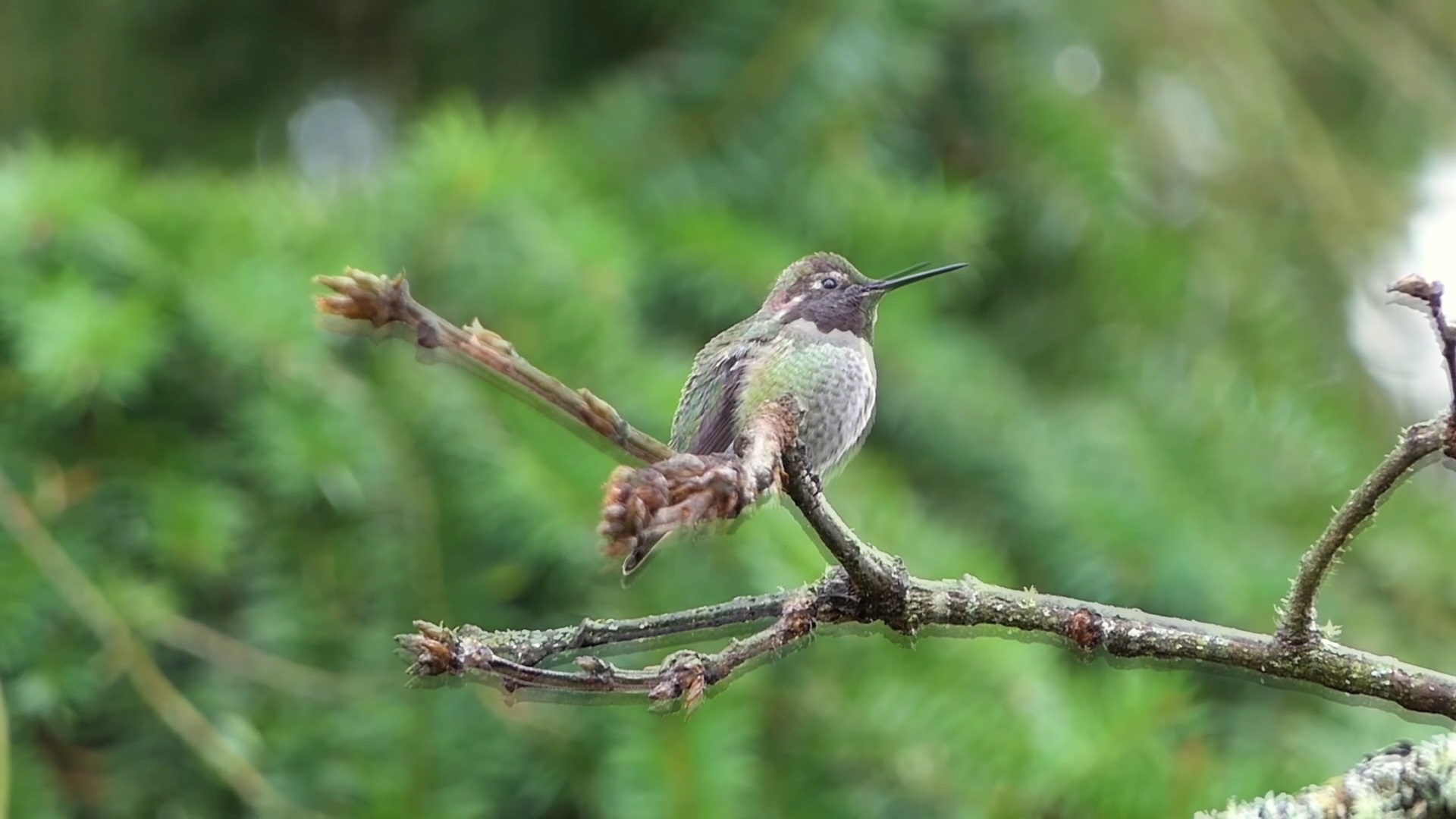}}
     \vspace{1pt}
     \centerline{\includegraphics[width=\textwidth]{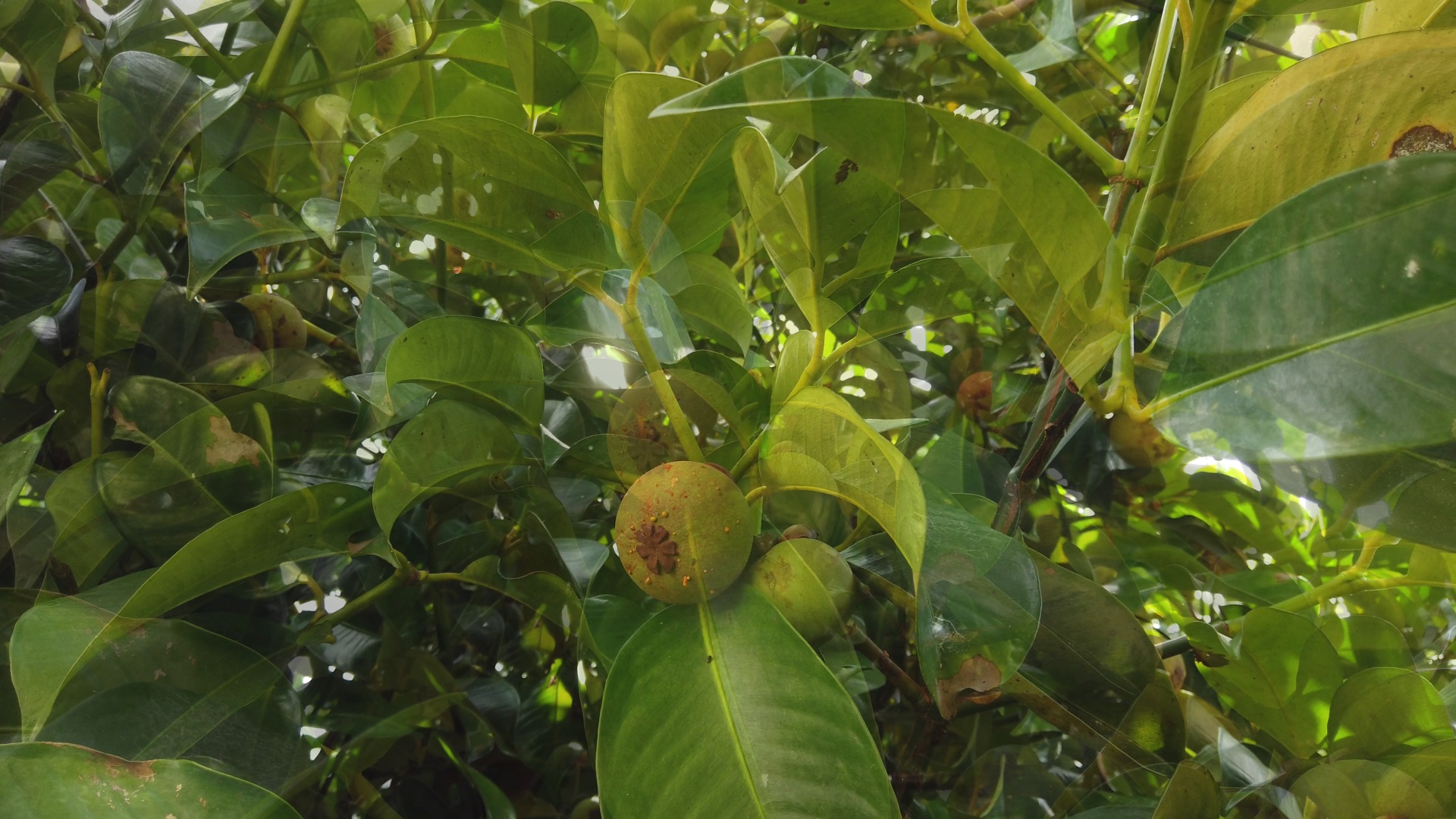}}
     \vspace{1pt}
     \centerline{\includegraphics[width=\textwidth]{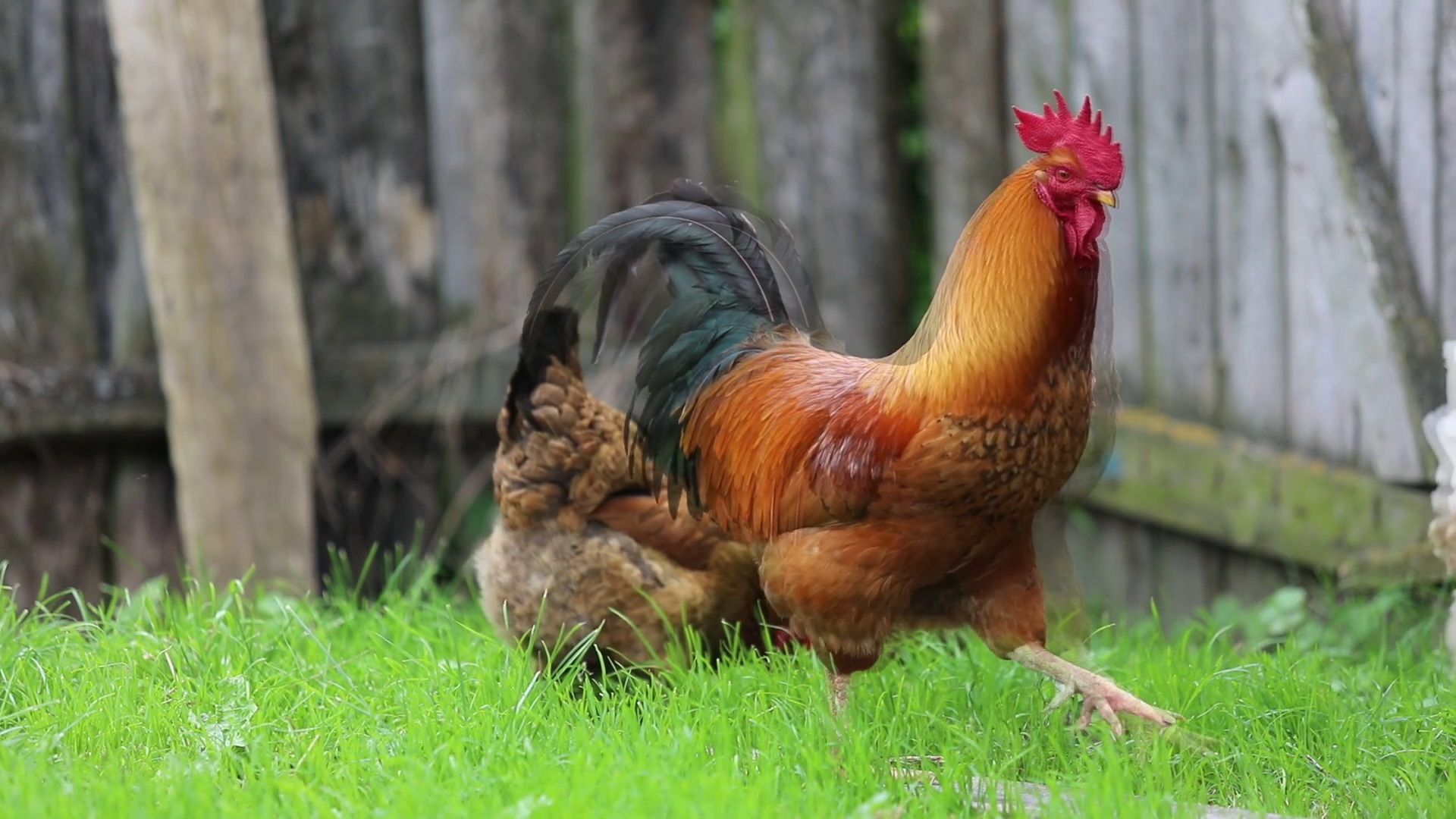}}
     \vspace{1pt}
     \centerline{Image Overlay}
 \end{minipage}
 \hfill
 \begin{minipage}{0.498\linewidth}
     \centerline{\includegraphics[width=\textwidth]{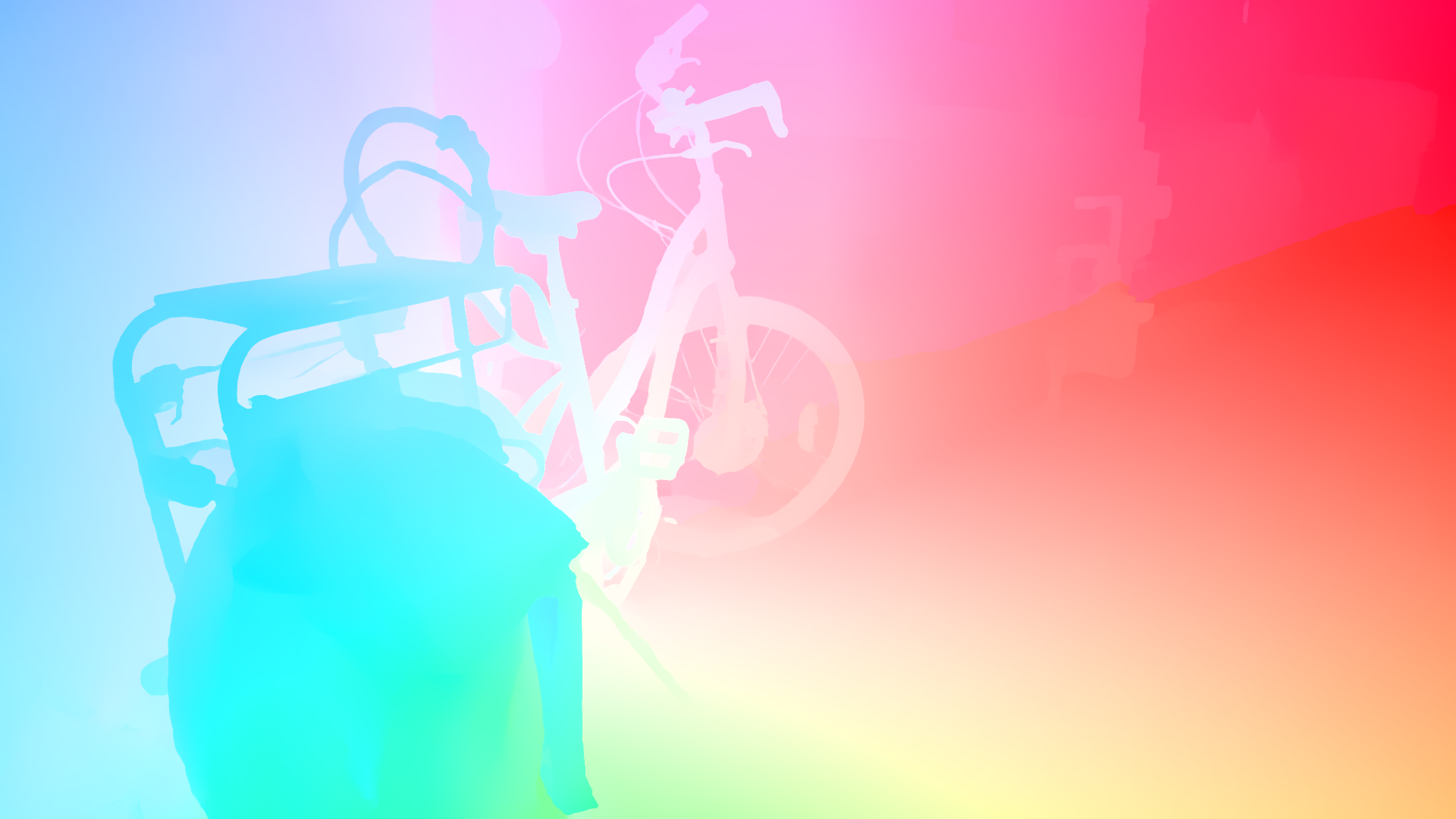}}
     \vspace{1pt}
     \centerline{\includegraphics[width=\textwidth]{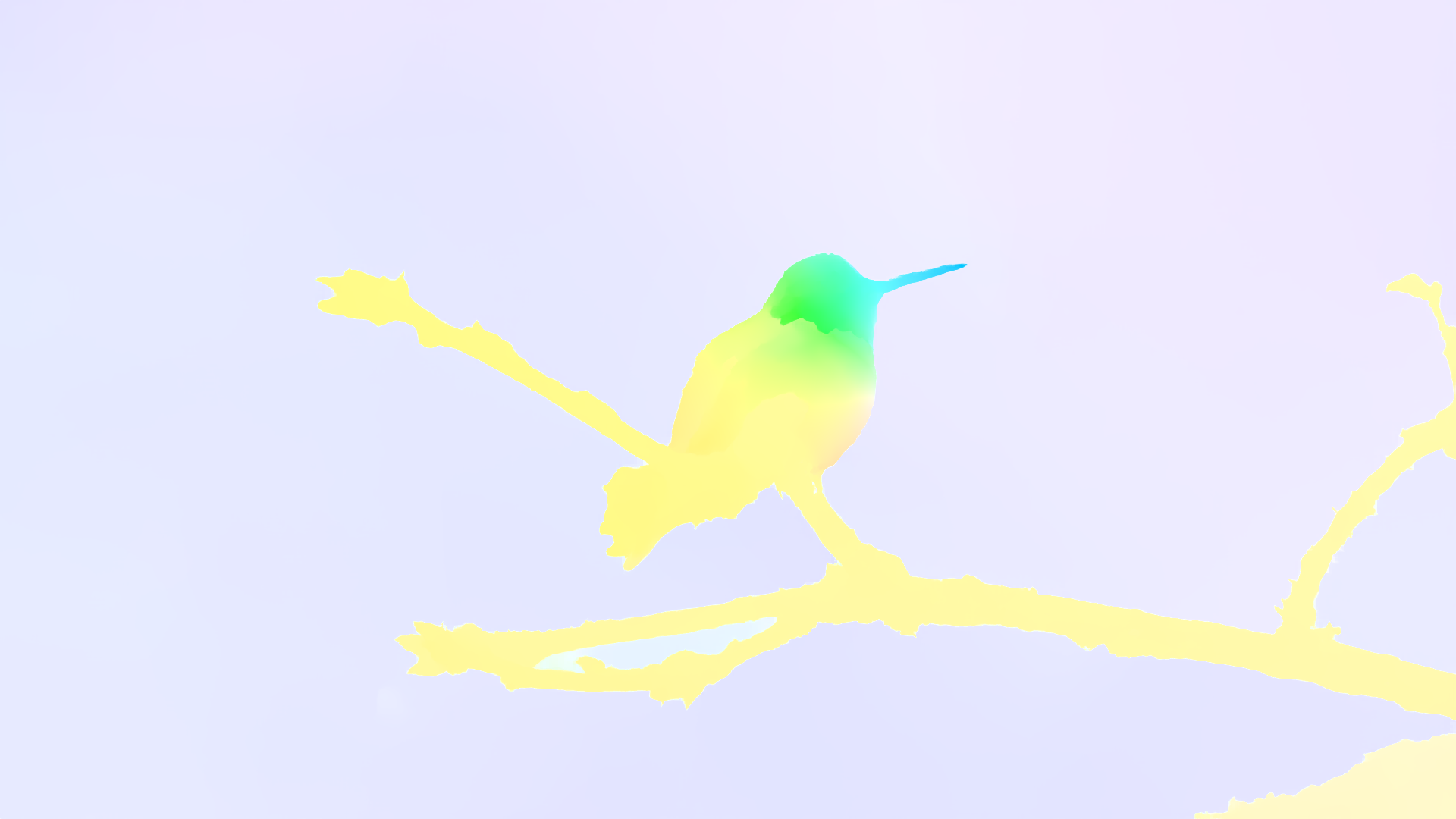}}
     \vspace{1pt}
     \centerline{\includegraphics[width=\textwidth]{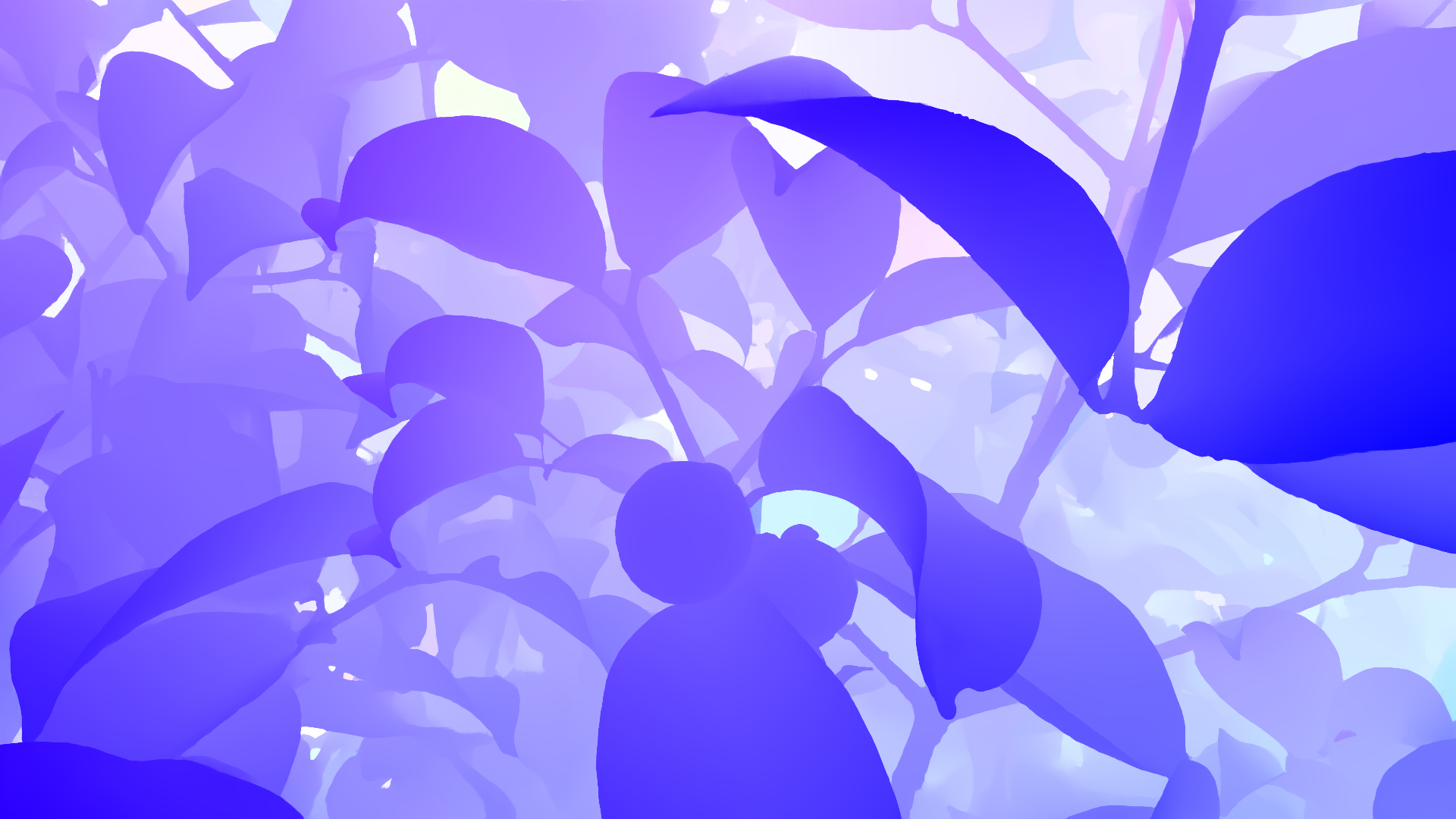}}
     \vspace{1pt}
     \centerline{\includegraphics[width=\textwidth]{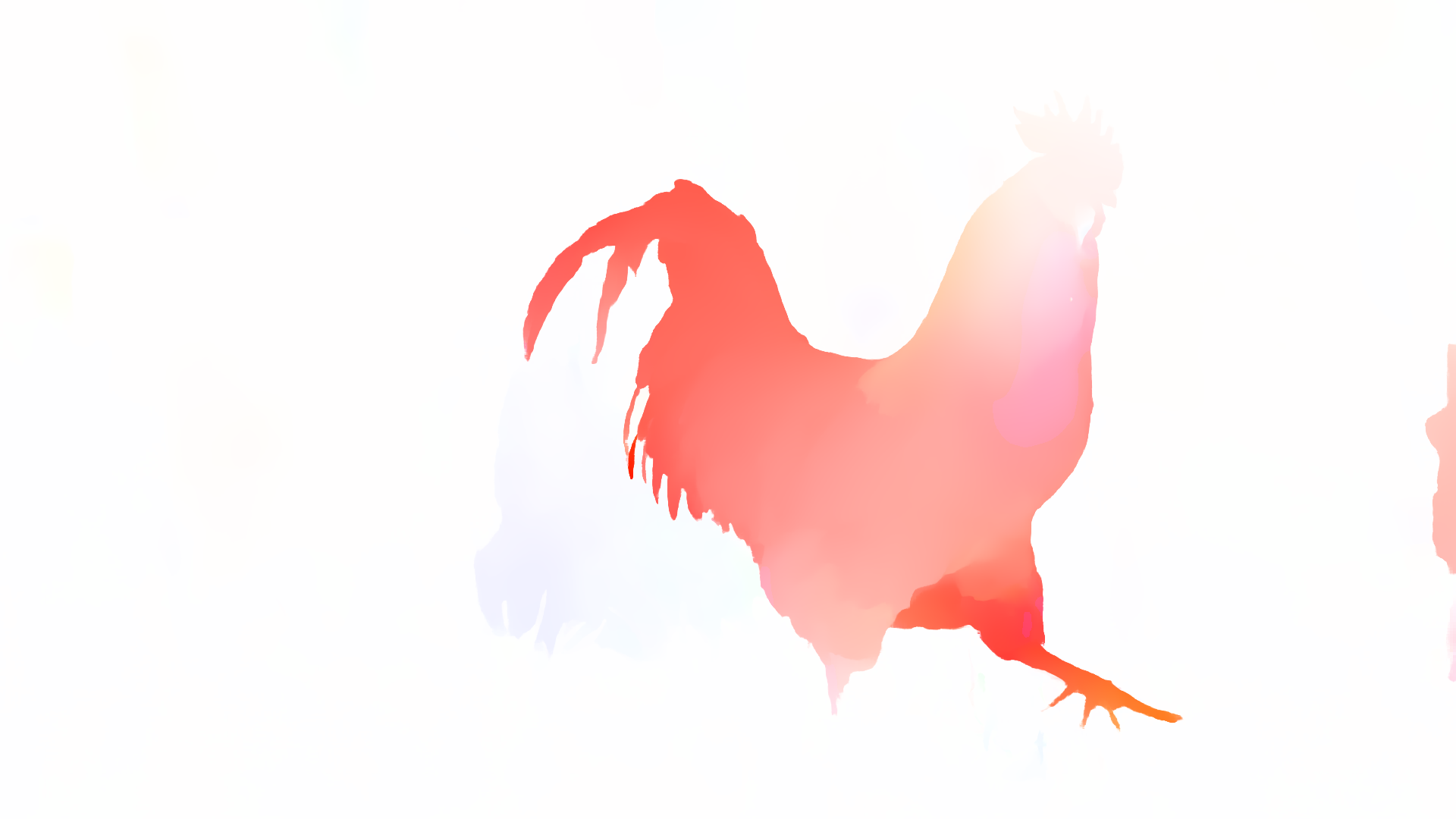}}
     \vspace{1pt}
     \centerline{Optical Flow}
 \end{minipage}
\caption{High-resolution optical flow results on public real-world images. The test resolution is $1080\times1920$ }
\label{fig:public_data}
\vspace{10pt}
\end{figure*}

\begin{figure*}[htbp]
 \begin{minipage}{0.498\linewidth}
     \vspace{1pt}
     \centerline{\includegraphics[width=\textwidth]{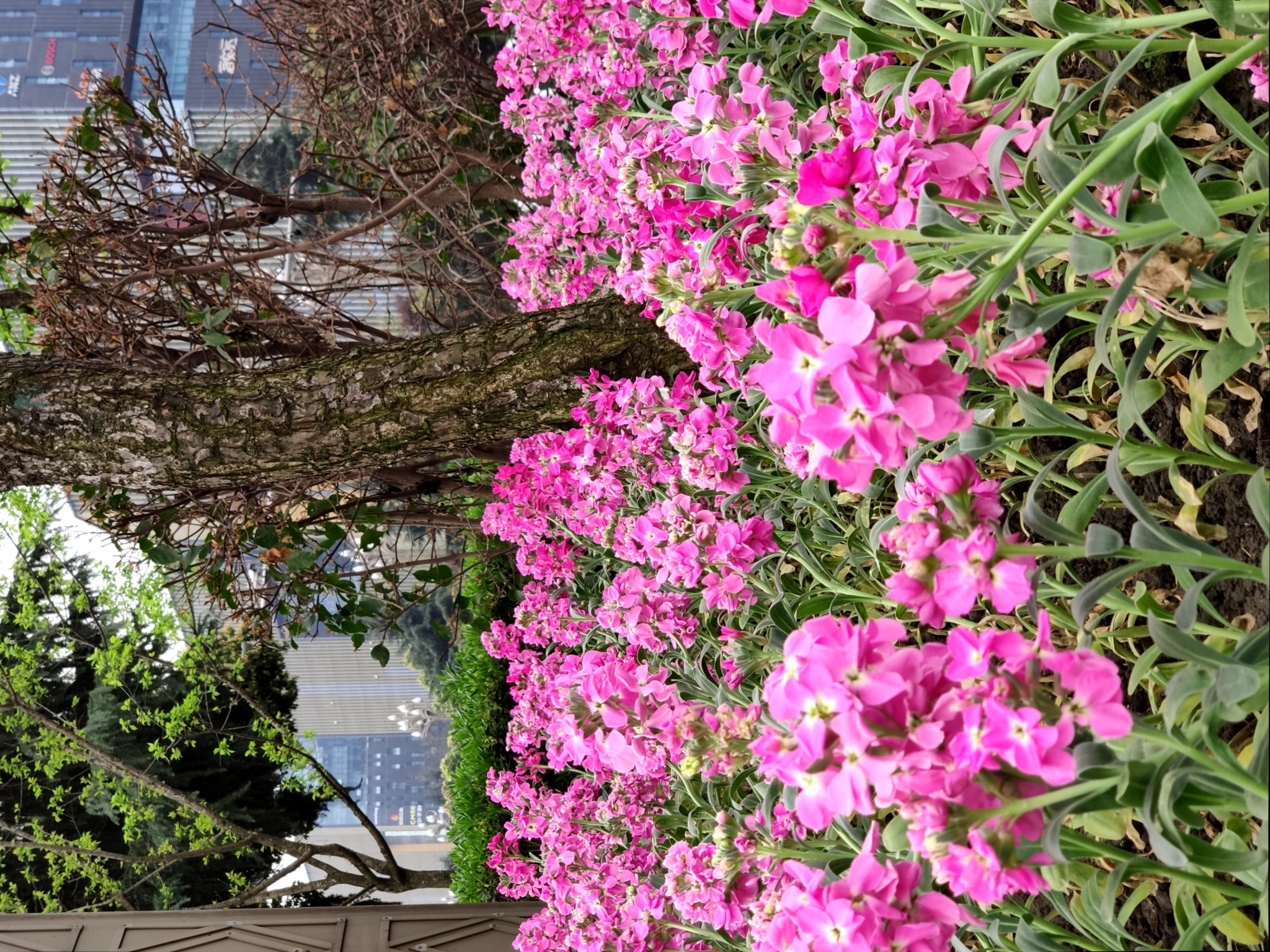}}
     \vspace{1pt}
     \centerline{\includegraphics[width=\textwidth]{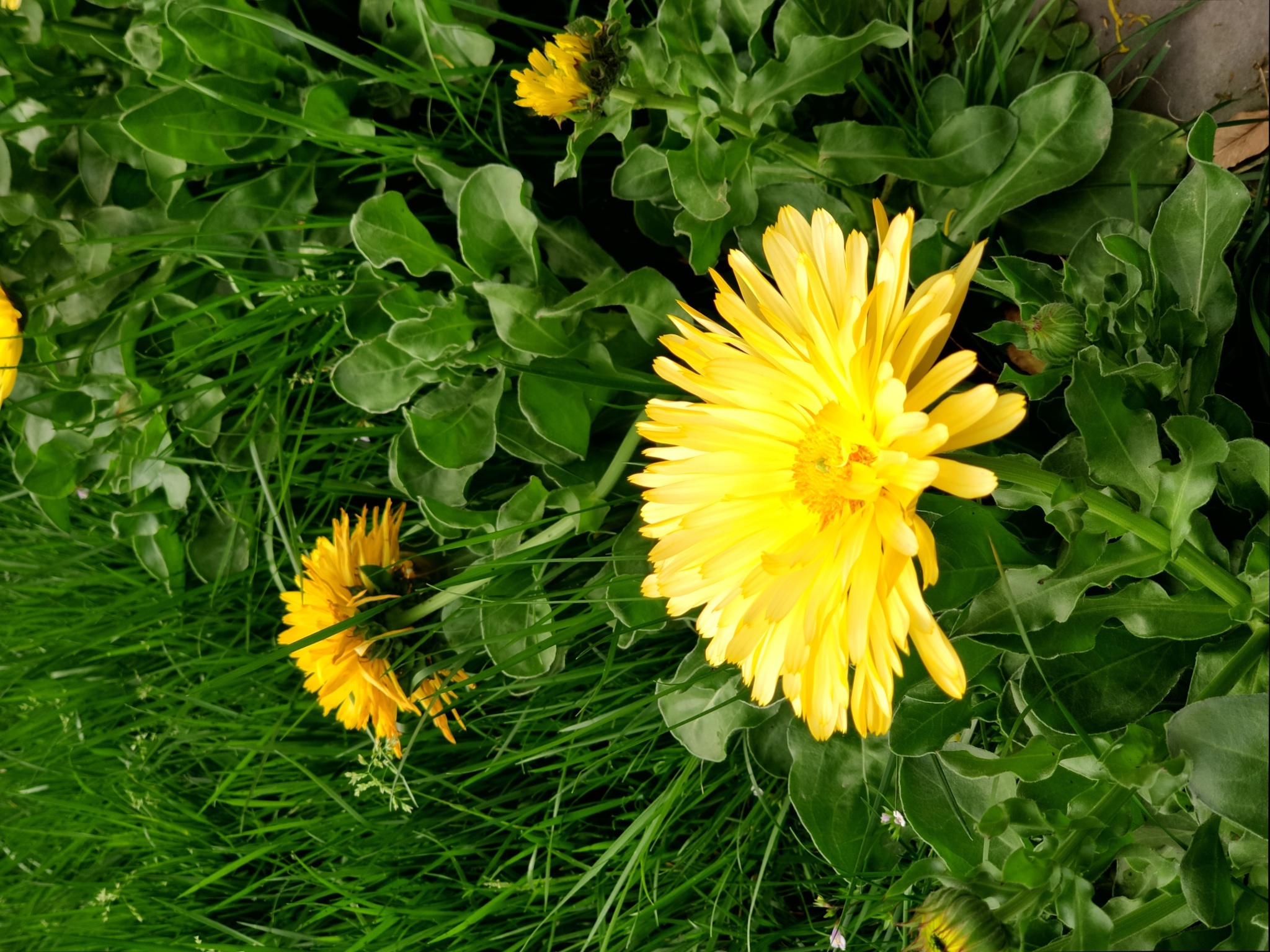}}
     \vspace{1pt}
     \centerline{\includegraphics[width=\textwidth]{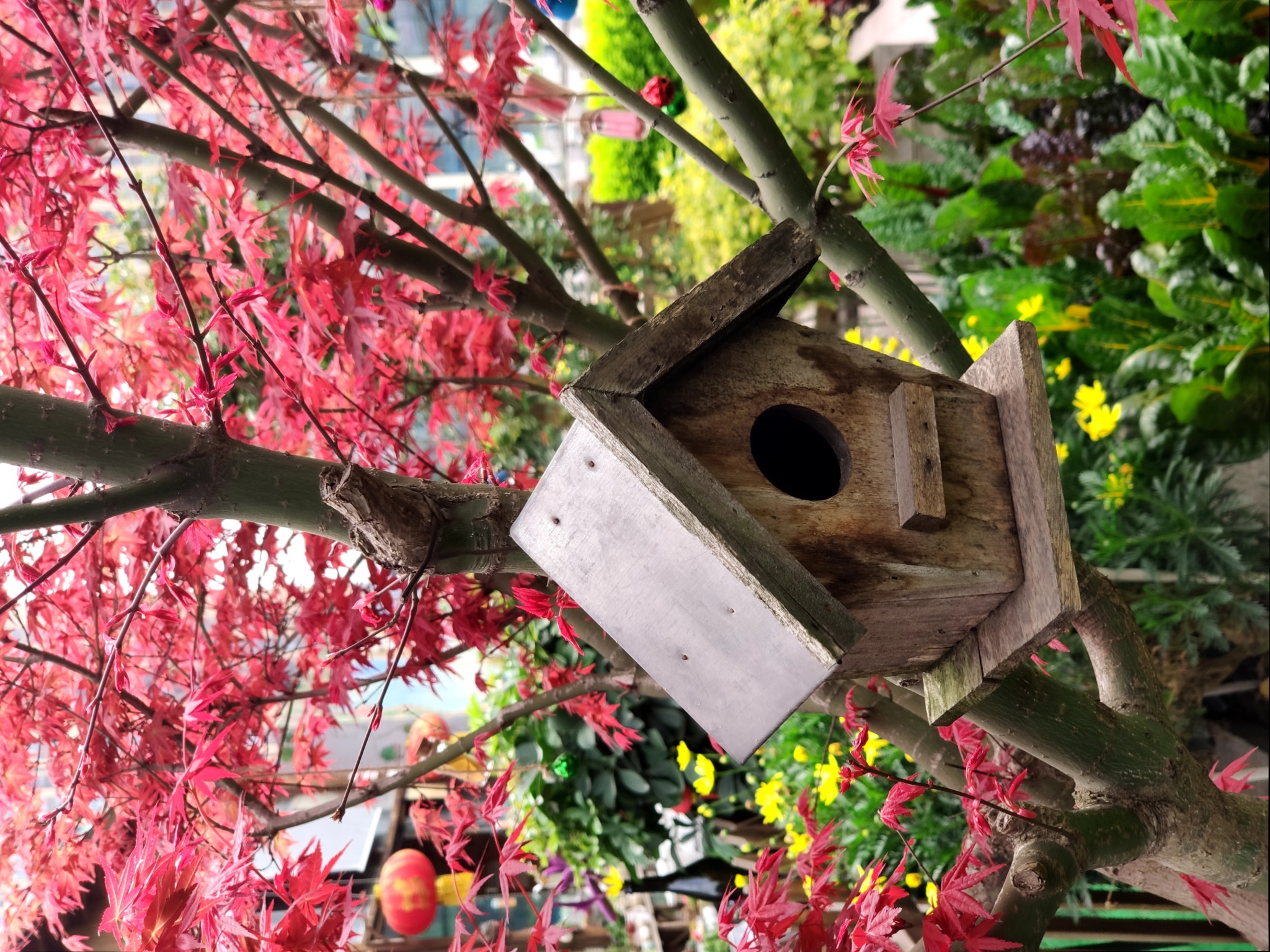}}
     \vspace{1pt}
     \centerline{Image1}
 \end{minipage}
 \hfill
 \begin{minipage}{0.498\linewidth}
    \centerline{\includegraphics[width=\textwidth]{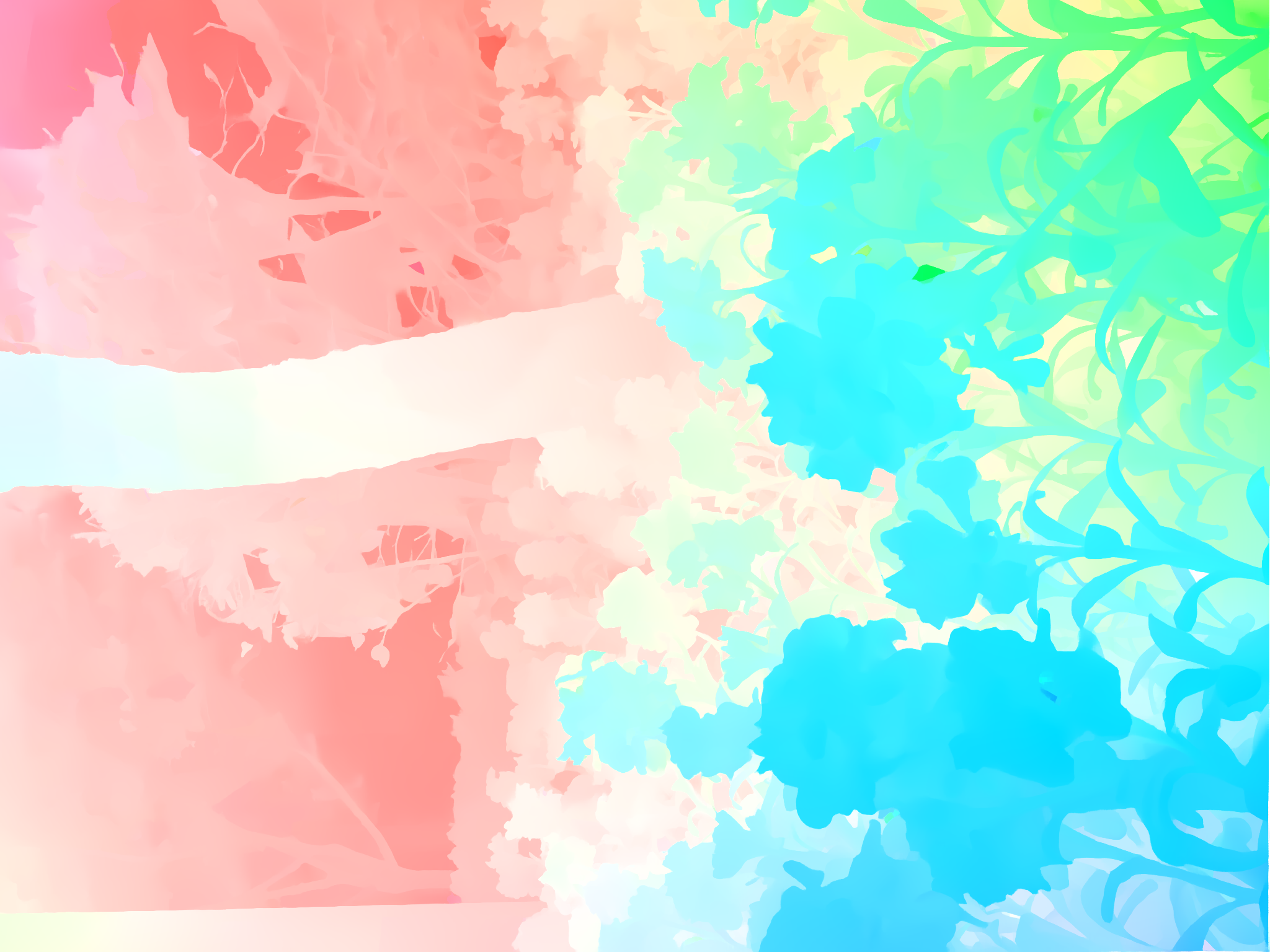}}
    \vspace{1pt}
    \centerline{\includegraphics[width=\textwidth]{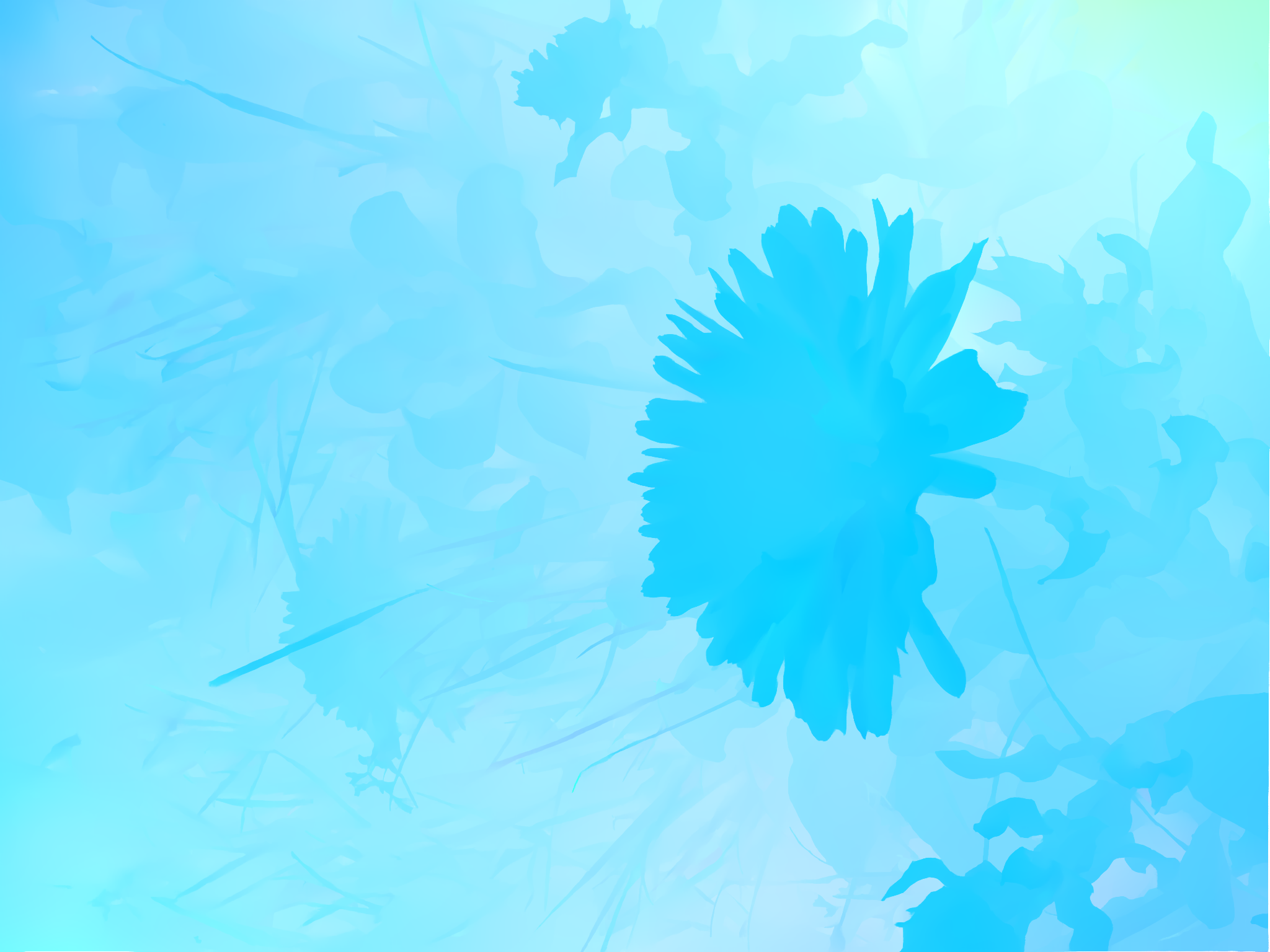}}
    \vspace{1pt}
    \centerline{\includegraphics[width=\textwidth]{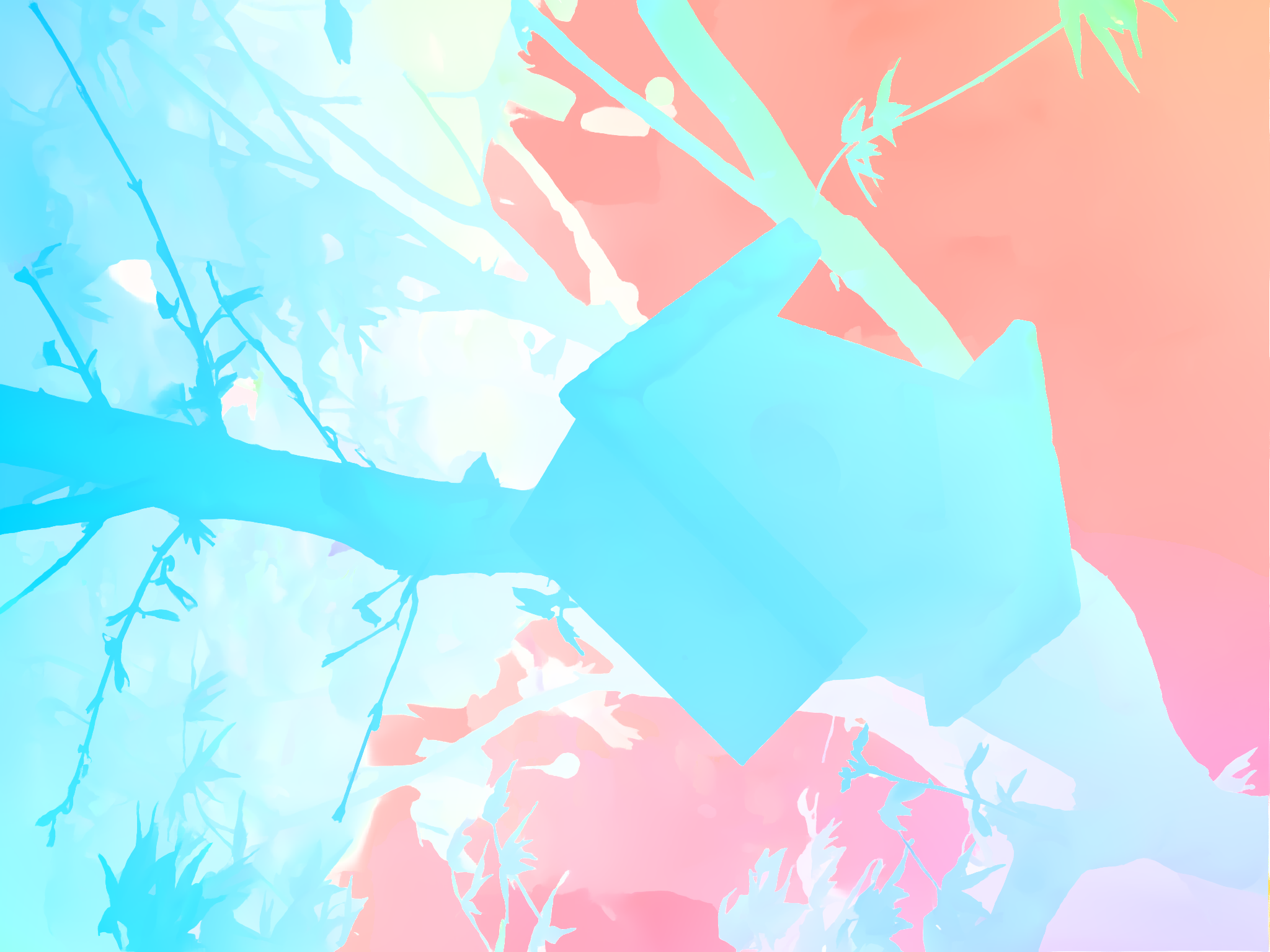}}
    \centerline{Optical Flow}
 \end{minipage}
\caption{High-resolution optical flow results on self-captured images. The test resolution is $1536\times2048$}
\label{fig:private_data}
\vspace{10pt}
\end{figure*}

To verify the robustness of optical flow in different high-resolution real-world scenes, 
we first tested DIP on the free used public dataset\footnote{\url{https://www.pexels.com/videos/}} with the resolution of $1080\times1920$ and showed results in \cref{fig:public_data}.
Then, we further used our mobile phone to collect images with a larger resolution($1536\times2048$) for testing and showed results in \cref{fig:private_data}. Experiments show that even if only virtual data is used for training, DIP still shows strong detail retention ability in high-resolution real-world scenes, which further confirms the strong cross-dataset generalization ability of DIP.

\section{Limitations}
\label{sec: DIP Limitations}
In the main paper, we observe that DIP is very friendly to the situations on fine-structure motions in the Sintel\cite{butler2012naturalistic} clean dataset (such as the person in the palace). However, a special weakness of our method is dealing with blurry regions, which is due to the limitations of neighborhood propagation of DIP. The entropy of the propagated information is greatly reduced when the features of the neighborhood are blurred, which leads to a weakening of the overall optical flow quality. An incorrect case is shown in \cref{fig:limit}. In the Sintel Clean images, DIP is able to estimate the optical flow that takes into account details and large displacement. However, in strong motion blur scenes of Sintel Final data, the propagation of incorrectly matched information in the neighborhood leads to incorrect predictions. In order to solve such problems, a non-local attention mechanism will be introduced in the further works.

\end{document}